\documentclass{article}

\usepackage{microtype}
\usepackage{graphicx}
\usepackage{subfigure}
\usepackage{booktabs} 

\usepackage{hyperref}

\usepackage[accepted]{icml2024}

\usepackage{amsmath}
\usepackage{amssymb}
\usepackage{mathtools}
\usepackage{amsthm}

\usepackage{multirow}

\usepackage[capitalize,noabbrev]{cleveref}

\theoremstyle{plain}
\newtheorem{theorem}{Theorem}[section]

\theoremstyle{definition}

\theoremstyle{remark}

\newcommand{\modelname}{HypDiff}
\usepackage[textsize=tiny]{todonotes}

\icmltitlerunning{Hyperbolic Geometric Latent Diffusion Model for Graph Generation}

\begin{document}

\twocolumn[
\icmltitle{Hyperbolic Geometric Latent Diffusion Model for Graph Generation}



\icmlsetsymbol{equal}{*}

\begin{icmlauthorlist}
\icmlauthor{Xingcheng Fu}{1}
\icmlauthor{Yisen Gao}{4}
\icmlauthor{Yuecen Wei}{3}
\icmlauthor{Qingyun Sun}{2}
\icmlauthor{Hao Peng}{2}
\icmlauthor{Jianxin Li}{2}
\icmlauthor{Xianxian Li}{1}
\end{icmlauthorlist}

\icmlaffiliation{1}{Key Lab of Education Blockchain and Intelligent Technology, Ministry of Education, Guangxi Normal University, Guilin, China}
\icmlaffiliation{2}{Beijing Advanced Innovation Center for Big Data and Brain Computing, School of Computer Science and Engineering, Beihang University, Beijing, China}
\icmlaffiliation{3}{School of Software, Beihang University, Beijing, China}
\icmlaffiliation{4}{Institute of Artificial Intelligence, Beihang University, Beijing, China}

\icmlcorrespondingauthor{Xingcheng Fu}{fuxc@gxnu.edu.cn}
\icmlcorrespondingauthor{Jianxin Li}{lijx@act.buaa.edu.cn}
\icmlcorrespondingauthor{Xianxian Li}{lixx@gxnu.edu.cn}

\icmlkeywords{Graph learning, latent diffusion model, hyperbolic space, graph generation}

\vskip 0.3in
]



\printAffiliationsAndNotice{}  

\begin{abstract}
Diffusion models have made significant contributions to computer vision, sparking a growing interest in the community recently regarding the application of them to graph generation. 
Existing discrete graph diffusion models exhibit heightened computational complexity and diminished training efficiency.
A preferable and natural way is to directly diffuse the graph within the latent space. 
However, due to the non-Euclidean structure of graphs is not isotropic in the latent space, the existing latent diffusion models effectively make it difficult to capture and preserve the topological information of graphs. 
To address the above challenges, we propose a novel geometrically latent diffusion framework \modelname~.
Specifically, we first establish a geometrically latent space with interpretability measures based on hyperbolic geometry, to define anisotropic latent diffusion processes for graphs. 
Then, we propose a geometrically latent diffusion process that is constrained by both radial and angular geometric properties, thereby ensuring the preservation of the original topological properties in the generative graphs.
Extensive experimental results demonstrate the superior effectiveness of \modelname~for graph generation with various topologies. 
\end{abstract}

\section{Introduction}
Graphs in the real world contain variety and important of topologies, and these topological properties often reflect physical laws and growth patterns, such as rich-clubs, small-worlds, hierarchies, fractal structures, etc.
Traditional random graph models based on graph theory, such as Erdos-Renyi~\cite{erdHos1960ERGRAPH}, Watts-Strogatz~\cite{watts1998collective} and Barabasi-Albert~\cite{barabasi1999emergence}, etc., need artificial heuristics to build the algorithms for single nature topologies and lack the flexibility to model various complex graphs.
Therefore, many deep learning models have been developed for graph generation, such as Variational Graph Auto-Encoder (VGAE)~\cite{kipf2016variational}, Generative Adversarial Networks(GAN)~\cite{goodfellow2014generative}, and other technologies. 
Recently, the \textit{Denoising Diffusion Probabilistic Model}(DDPM)~\cite{ho2020ddpm} have demonstrated great power and potential in image generation, attracting huge attention from the community of graph learning. 

\begin{figure*}[!t]
\centering
\subfigure[Original structure.]{\label{fig:structure}
\includegraphics[height=0.145\textheight]{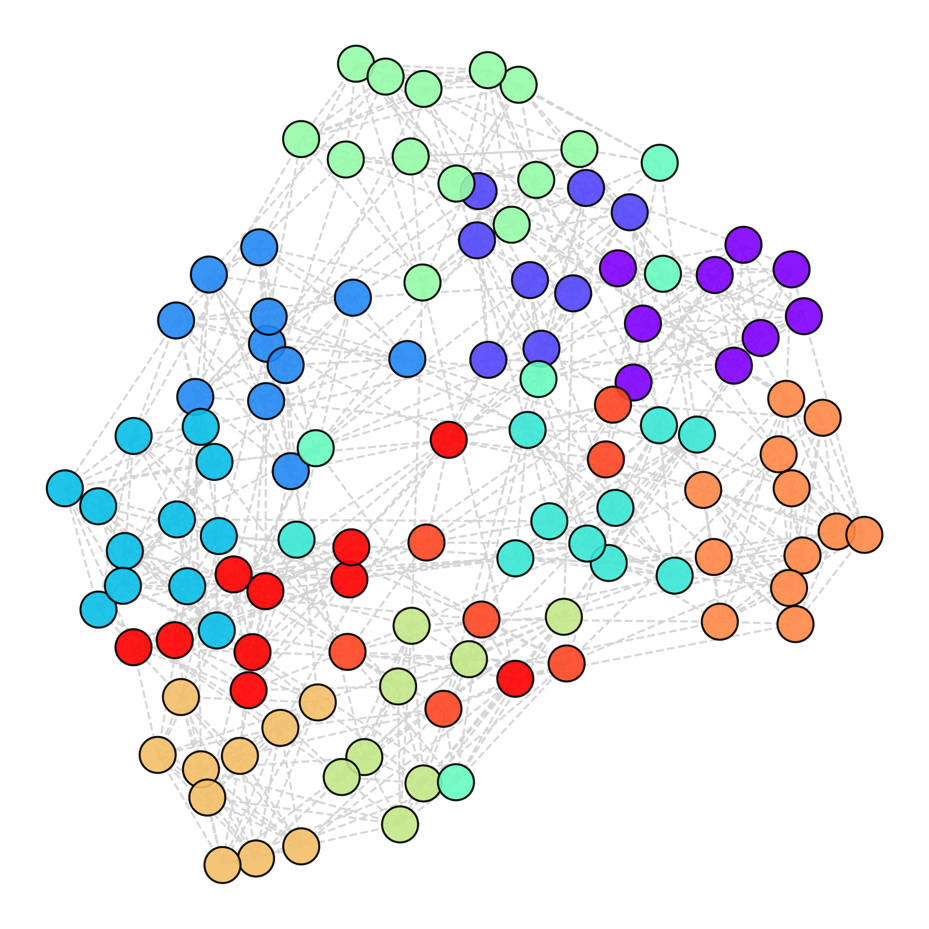}
}
\subfigure[Euclidean latent space.]{\label{fig:euclidean}
\includegraphics[height=0.145\textheight]{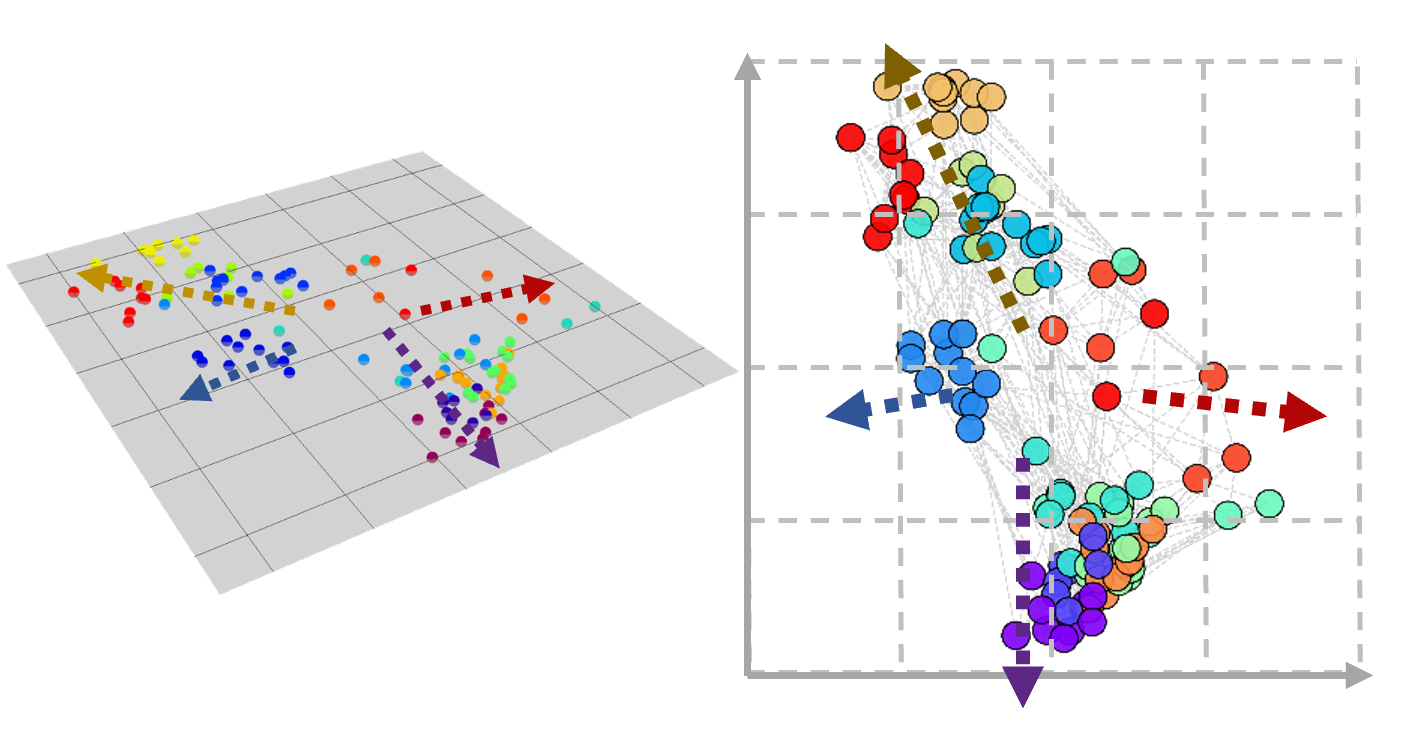}
}
\subfigure[Hyperbolic latent space.]{\label{fig:hyperbolic}
\includegraphics[height=0.19\linewidth]{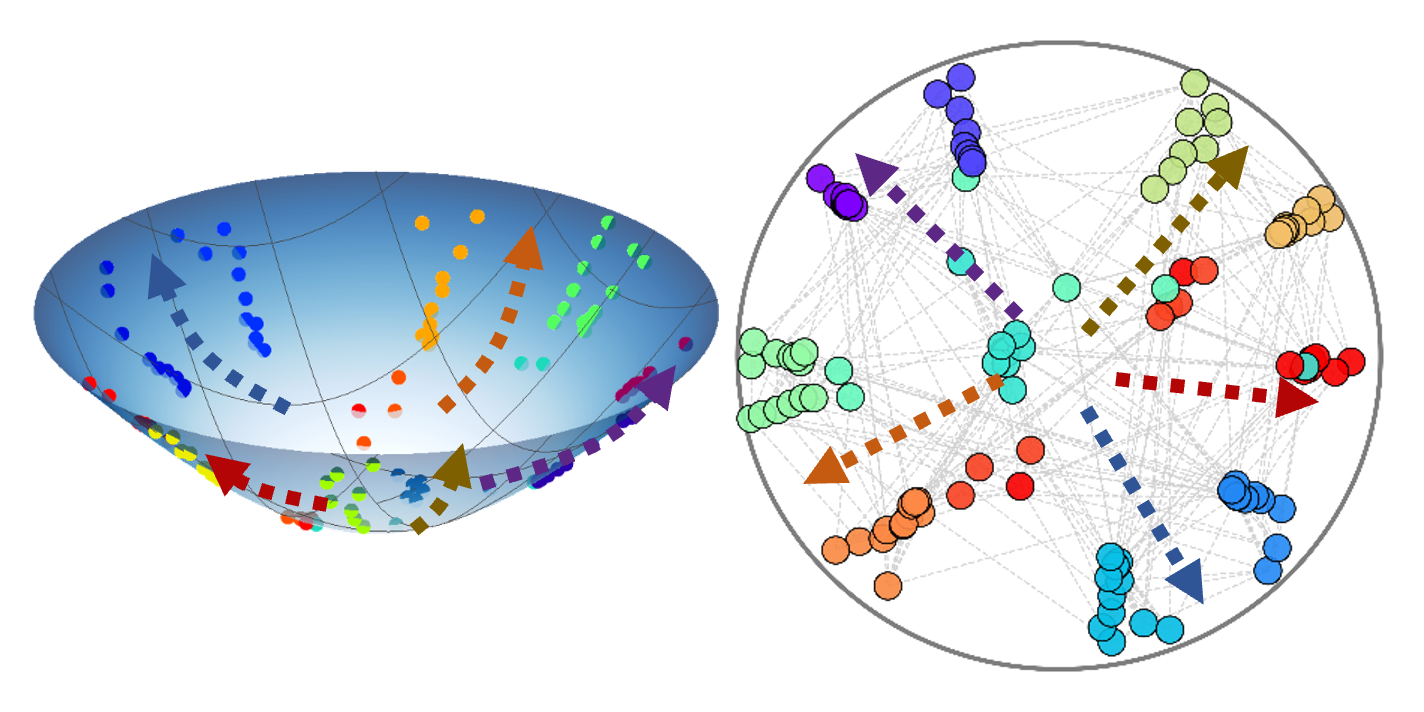}
}
\vspace{-1em}
\caption{
Visualization of node embeddings by singular value decomposition (SVD);
(a) Original structure visualization of the NCAA football graph and different colors indicate different labels(teams); 
(b) Visualization of node embeddings in 2D Euclidean space and planar projection;
(c) Visualization of node embeddings in 2D hyperbolic space and Poincar\'e disk projection.}

\label{fig:PSO}
\end{figure*}

For graph generation, a straightforward idea involves designing discretized diffusion methods for the graph structural information.~\cite{vignac2022digress,jo2022GDSS,luo2022GSDM}, and the other way is to develop advanced graph encoders to preserve structural information throughout the diffusion process within a continuous potential space~\cite{xu2021geodiff,xu2023GEOLDM}.
However, because of the irregular and non-Euclidean structure of graph data, the realization of the diffusion model for graphs still has two main limitations: 
\textbf{(1) High computational complexity.} 
The core to graph generation is to handle the discreteness, sparsity and other topological properties of the non-Euclidean structure.
Since the Gaussian noise perturbation used in the vanilla diffusion model is not suitable for discrete data, the discrete graph diffusion model usually has high time and space complexity due to the problem of structural sparsity. 
Moreover, the discrete graph diffusion model relies on a continuous Gaussian noise process to create fully connected, noisy graphs~\cite{zhang2023survey,ingraham2019generativeprotein} which loses structural information and underlying topological properties. 
\textbf{(2) Anisotropy of non-Euclidean structure.} 
Different from the regular structure data (e.g. pixel matrix or grid structure), the "irregular" non-Euclidean structure embeddings of graph data are anisotropic in continuous latent space~\cite{elhag2022GraphAnisotropic}. 
As shown in Figure~\ref{fig:euclidean}, the node embeddings of a graph in Euclidean space exhibit significant anisotropy in several specific directions.
Recently, some studies~\cite{yang2023directional} have shown that isotropic diffusion of the node embedding of the graph in the latent space will treat the anisotropic structural information as noise, and this useful structural information will be lost in the denoising process.

\begin{figure*}[ht]
\centering
\subfigure[Geometric interpretation.]{\label{fig:GI}
\includegraphics[height=0.30\textheight]{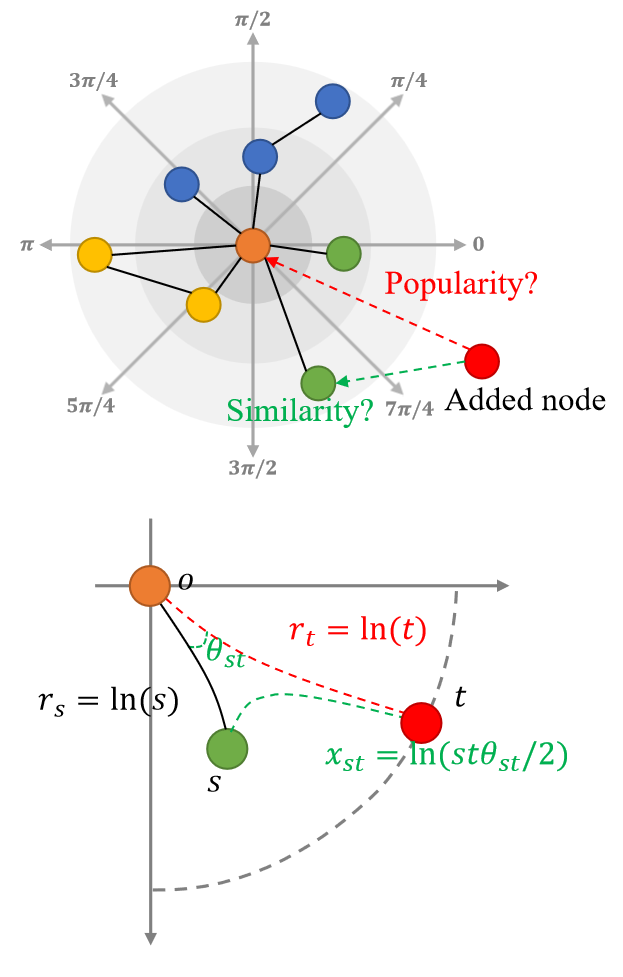}
}
\subfigure[Hyperbolic latent diffusion.]{\label{fig:diff}
\includegraphics[height=0.38\linewidth]{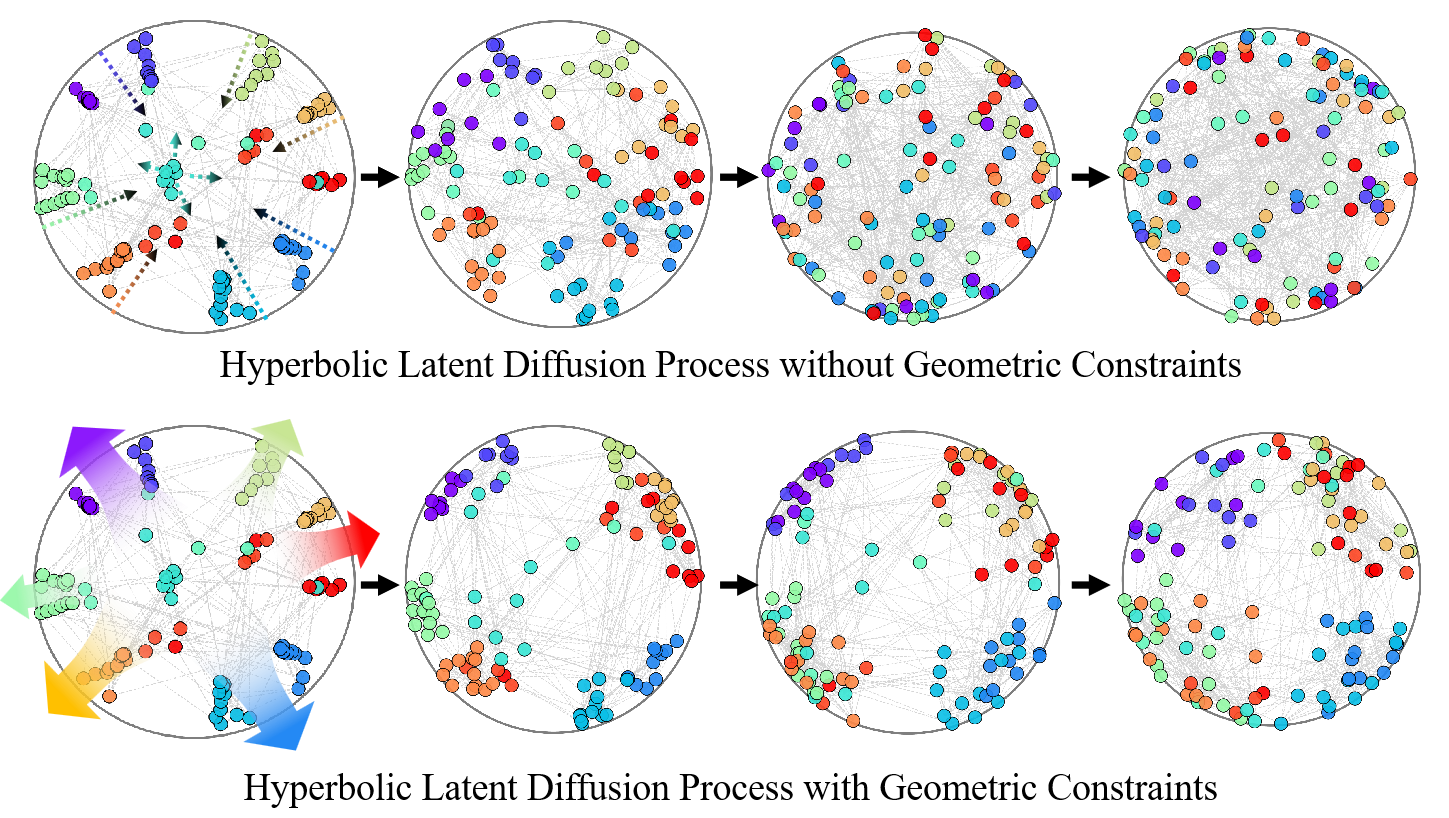}
}
\centering
\vspace{-1em}
\caption{(a) Geometric interpretation of the hyperbolic geometry, which unifies the radius and angle measurements in polar coordinates and interprets as popularity and similarity respectively;
(b) Hyperbolic latent diffusion processing with isotropic/anisotropic noise;}
\label{fig:GEO}
\end{figure*}

Hyperbolic geometric space is widely recognized as an ideal continuous manifold for representing discrete tree-like or hierarchical structures~\cite{cannon1997hyperbolic,ungar1999hyperbolic, Krioukov2010Hyperbolic, SunAAAI24}, and has been widely studied and applied to various graph learning tasks~\cite{sun2021hyperbolic, PoincareGlove, NickelK17Poincare, sala2018representation, HGCN_ChamiYRL19, SunWWW24}. 
Inspired by these studies, we find that hyperbolic geometry has great potential to address non-Euclidean structural anisotropy in graph latent diffusion processes.
As shown in Figure~\ref{fig:hyperbolic}, in hyperbolic space, we can observe that the distribution of node embeddings tends to be isotropic globally, while anisotropy is preserved locally.
In addition, hyperbolic geometry unifies angular and radial measures of polar coordinates as shown in Figure~\ref{fig:GI}, and can provide geometric measures with physical semantics and interpretability~\cite{papadopoulos2012popularity}. 
It is exciting that hyperbolic geometry can provide a geometrically latent space with graph geometric priors, able to help deal with the anisotropy of graph structures by special geometric measures. 

Based on the above insights, we aim to establish a suitable geometrically latent space based on hyperbolic geometry to design an efficient diffusion process to the non-Euclidean structure for topology-preserving graph generation tasks. 
However, there are two primary challenges: (1) the additivity of continuous Gaussian distributions is undefined in hyperbolic latent space; (2) devising an effective anisotropic diffusion process for non-Euclidean structures.

\textbf{Contributions.}
To address the challenges, we propose a novel \underline{\textbf{Hyp}}erbolic Geometric Latent \underline{\textbf{Diff}}usion (\textbf{\modelname}) model for the graph generation.  
For the additive issue of continuous Gaussian distribution in hyperbolic space, we propose an approximate diffusion process based on radial measures.
Then the angular constraint was utilized to constrain the anisotropic noise to preserve more structural prior, guiding the diffusion model to finer details of the graph structure.
Our contributions are summarized as: 

\begin{itemize}
\item 
We are the first to study the anisotropy of non-Euclidean structures for graph latent diffusion models from a geometric perspective, and propose a novel hyperbolic geometric latent diffusion model~\modelname.
\item 
We proposed a novel geometrically latent diffusion process based on radial and angular geometric constraints in hyperbolic space, and addresses the additivity of continuous Gaussian distributions and the issue of anisotropic noise addition in hyperbolic space. 
\item 
Extensive experiments on synthetic and real-world datasets demonstrate a significant and consistent improvement of \modelname~and provide insightful analysis for graph generation. 
\end{itemize}

\section{Related Works}
\subsection{Graph Generative Diffusion Model}
Different from that learn to generate samples once, like GAN~\cite{goodfellow2014generative,wang2018graphgan,dai2018adversarial}, VGAE~\cite{yu2018learning,xu2018spherical,grattarola2019adversarial} or GraphRNN~\cite{GraphRNN2018}, the diffusion model~\cite{ho2020ddpm} aims to gradually convert the sample into pure noise by a parameterized Markov chain process. 
Some recent works~\cite{xu2021geodiff,xu2023GEOLDM} employ advanced graph encoders to effectively preserve the inherent structural information throughout the diffusion process within a continuous potential space.
Gaussian noise is added on the distribution of nodes and edges of the graph~\cite{vignac2022digress}, and Gaussian processes are performed on the neighborhood or spectral domain of the graph~\cite{vignac2022digress,jo2022GDSS,luo2022GSDM}.
However, existing discrete diffusion models have many challenges in capturing the non-Euclidean structure and preserving underlying topological properties.

\subsection{Hyperbolic Graph Learning}
Hyperbolic geometric space was introduced into complex networks earlier to represent the small-world and scale-free complex networks~\cite{Krioukov2010Hyperbolic,papadopoulos2012popularity}. 
With high capacity and hierarchical-structure-preserving ability, hyperbolic geometry is also used in NLP~\cite{NickelK17Poincare, PoincareGlove} to learn word representations with hypernym structure. 
For graph neural networks, hyperbolic space is recently introduced into graph neural networks~\cite{HGNN_Qi, HGCN_ChamiYRL19, sun2021hyperbolic, SunAAAI22}. 
$\mathcal{P}$-VAE~\cite{mathieu2019poincare} and Hyper-ANE~\cite{HyperANE} extend VAE and GAN into the hyperbolic versions to learn the hierarchical representations. 
To sum up, hyperbolic geometry provides an intuitive and efficient way of understanding the underlying structural properties of the graph.

\begin{figure*}[htb]
\centering
\includegraphics[width=1\textwidth]{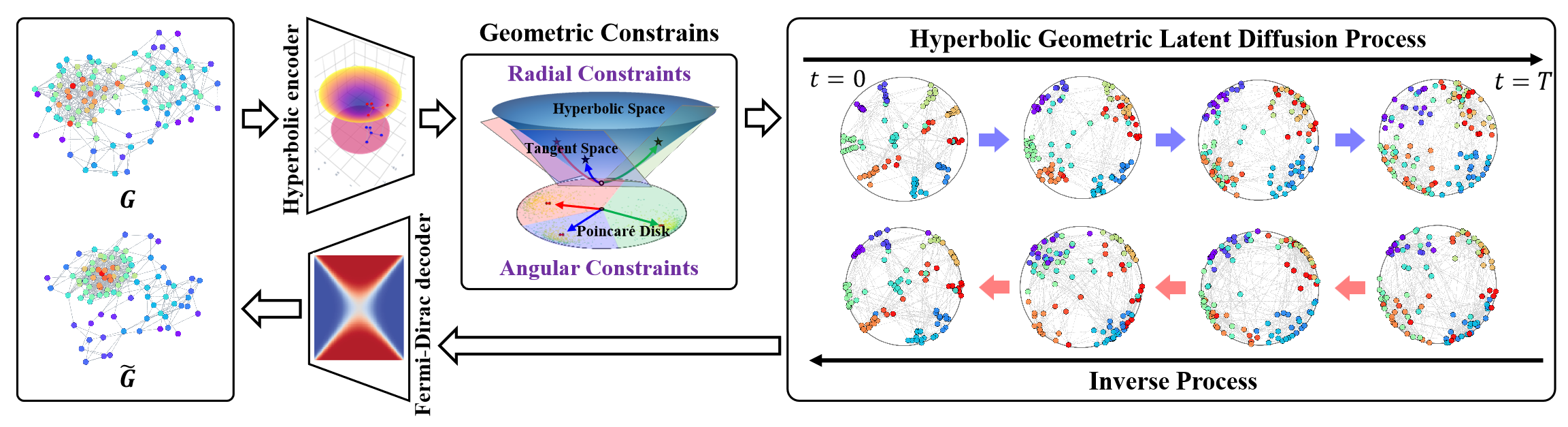}
\vspace{-1em}
\caption{An illustration of \modelname~ architecture.} 
\label{fig:Architecture}
\end{figure*}

\section{Methodology}
In this section, we present our \textbf{Hyp}erbolic geometric latent \textbf{Diff}usion model (\textbf{\modelname}) for addressing the two main challenges. 
The key insight is that we leverage hyperbolic geometry to abstract the implicit hierarchy of nodes in the graph and introduce two geometric constraints to preserve important topological proprieties, such as scale-free, navigability, and modularity. 
Considering the successful experiences of graph latent diffusion models~\cite{xu2023GEOLDM}, we adopt a two-stage training strategy framework in our practice.
We first train the hyperbolic autoencoder to obtain the pre-trained node embeddings, and then train the hyperbolic geometric latent diffusion process.
The architecture is shown in Figure~\ref{fig:Architecture}.

\subsection{Hyperbolic Geometric Autoencoding}
We first need to embed the graph data $\mathcal{G}=(\mathbf{X},\mathbf{A})$ into a low-dimensional hyperbolic geometric space to improve the graph latent diffusion process. 

\noindent \textbf{Hyperbolic Encoder and Decoder.}
We consider a hyperbolic variant of the auto-encoder, consisting of the \textit{hyperbolic geometric encoder} and the \textit{Fermi-Dirac decoder}.
Where the \textit{hyperbolic geometric encoder} encodes the graph $\mathcal{G}=(\mathbf{X},\mathbf{A})$ into the hyperbolic geometric space to obtain a suitable hyperbolic representation, and the \textit{Fermi-Dirac decoder }decodes the hyperbolic representation back into the graph data domain.
The hyperbolic manifold $\mathbb{H}^d$ and the tangent space $\mathcal{T}_{\mathrm{x}}$ can be mapped to each other via \textit{exponential map} and \textit{logarithmic map}~\cite{Octavian2018HyperbolicNeuralNetworks}. 
Then, we can leverage Multi-Layer Perceptrons(MLP) or Graph Neural Networks(GNNs) by exponential and logarithmic mapping as hyperbolic geometric encoders.
In this paper, we use Hyperbolic Graph Convolutional Neural Networks(HGCN)~\cite{HGCN_ChamiYRL19} as the hyperbolic geometric encoder.

\noindent \textbf{Optimization of Autoencoding.}
Due to the additive failure of the Gaussian distribution in hyperbolic space, we cannot directly use \textit{Riemannian normal distribution} or \textit{wrapped normal distribution}. 
Instead of hyperbolic diffusion embedding~\cite{HDE} using the product space of multiple manifolds, we propose a new diffusion process in hyperbolic space, which will be described in detail in Section~\ref{subsec:hypdiff}.
Following $\mathcal{P}$-VAE~\cite{mathieu2019poincare}, for compute efficiency, the Gaussian distribution of hyperbolic space is approximated by the Gaussian distribution of the tangent plane $\mathcal{T}_{\mu}$.
The optimization of hyperbolic geometric auto-encoding is as follows:
\begin{equation}\label{Eq:encoding}
\begin{aligned}
    \mathcal{L}_{\mathrm{HAE}} = -\mathbb{E}_{q_\phi(\mathbf{z}_{\mathbf{x}}|\mathbf{x})} \mathrm{logmap}_{\mathbf{o}}^{c} p_\xi
    \left(\mathbf{x}|\mathbf{z}_{\mathbf{x}}\right),
\end{aligned}
\end{equation}
where $\log_{\mathbf{o}}^{c}$ is the logarithmic mapping of the north pole (origin) $\mathbf{o}$ of hyperbolic space to simplify the computation. 

\subsection{Hyperbolic Geometric Latent Diffusion Process}\label{subsec:hypdiff}
Unlike the linear addition in Euclidean space, hyperbolic space utilizes M\"{o}bius addition, posing challenges for diffusion over a hyperbolic manifold.
Furthermore, the isotropic noise leads to a rapid reduction of signal-to-noise ratio making it difficult to preserve topological information, and for the detailed results and analysis please refer to Appendix~\ref{app:snr}. 
In light of these issues, we propose a novel diffusion process to address both of them.

\textbf{Hyperbolic Anisotropic Diffusion.}
The anisotropy of the graph in the latent space contains an inductive bias of the graph structure, where the most critical challenge is how to determine the dominant directions of the anisotropic features. 
In additionally, on hyperbolic manifolds,  neither the wrapped normal distribution of the isotropic setup nor the anisotropic setup satisfies this property:
\begin{equation}
\begin{aligned}
&\boldsymbol{\eta} \not \sim \boldsymbol{\eta}_1 \oplus _{c}\boldsymbol{\eta}_2 ,\\
&\boldsymbol{\eta} \sim \mathcal{N}^{c}_{\mathbb{H}}\left(0, (\sigma_{1}^2+\sigma_{2}^2) I\right) ,\\
&\boldsymbol{\eta}_1 \sim \mathcal{N}^{c}_{\mathbb{H}}\left(0, \sigma_{1}^2 I\right), \boldsymbol{\eta}_2 \sim\mathcal{N}^{c}_{\mathbb{H}}\left(0, \sigma_{2}^2 I\right).
\end{aligned}
\end{equation}
where $c$ is Hyperbolic curvature and $ \mathcal{N}^{c}_{\mathbb{H}}$ is the Wrapped Gaussian distribution.
We propose a hyperbolic anisotropic diffusion framework to solve both challenges. The detailed proof process can be found in the Appendix \ref{pro:add}.
The core idea is to select the main diffusion direction (i.e., angle) based on the similarity clustering of nodes, which is equivalent to dividing the hyperbolic latent space into multiple sectors. 
Then we project the nodes of each cluster onto its center's tangent plane for diffusion.

Let $\mathbf{h}$ denote the embedding of the graph in the hyperbolic space and $\mathbf{h}_i$ denote the $i$-th node in it. Let $\mathbf{h}_i$ belong to the $k$-th cluster and its clustering center coordinates are $\mathbf{\mu}_k$, then the node $\mathbf{h}_i$ is represented in the  tangent space of $\mathbf{\mu}_k$ as $x_{0_i}$:
\begin{equation}\label{eq:noisy}
\begin{aligned}
&\mathbf{x}_{0_i} = \mathrm{logmap}_{\mathbf{\mu}_{k}}^{c}\left( {h}_{i}\right).
\end{aligned}
\end{equation}
where $\mathbf{\mu}_{k}$ is the central point of cluster $k$ obtained by Hyperbolic-Kmeans (h-kmeans)~\cite{hajri:hkmeans} algorithm. 
Note that the clusters can be obtained by any clustering algorithm based on similarity in the pre-processing stage.
Moreover, the hyperbolic clustering parameter $k$ has the following property:
\begin{theorem}\label{thm:k}
Given the hyperbolic clustering parameter $k \in [1,n]$, which represents the number of sectors dividing the hyperbolic space (disk).
The hyperbolic anisotropic diffusion is equivalent to directional diffusion in the Klein model $\mathbb{K}^{n}_{c}$ with multi-curvature ${c_{i\in{|k|}}}$, which is an approximate projecting onto the tangent plane set $\mathcal{T}_{\mathbf{o}_{i\in\{|k|\}}}$ of the centroids $\mathbf{o}_{i\in\{|k|\}}$. 
\end{theorem}
The proof is in the Appendix~\ref{proof:k}. 
This property elegantly establishes the relationship between our approximation algorithm and the Klein model with multiple curvatures. 
Our algorithm exhibits specific behaviors based on the value of $k$, it allows for a more flexible and nuanced representation of anisotropy based on the underlying hyperbolic geometry, enabling improved accuracy and efficiency in subsequent noise addition and training.

\noindent \textbf{Geometric Constraints. } 
Hyperbolic geometry can naturally and geometrically describe the connection pattern of nodes during graph growth~\cite{papadopoulos2012popularity}. 
As shown in Figure~\ref{fig:GI}, the popularity of a node can be abstracted by its radial coordinates and the similarity can be expressed by its angular coordinate distances in the hyperbolic space, and more detail can be referred to Appendix~\ref{app:hyperbolic}.

Our goal is to model a diffusion with geometric radial growth, and where this radial growth is consistent with hyperbolic properties. Considering that we need to maintain this kind of hyperbolic growth tendency in the tangent plane, we use the following formulas:
\begin{equation}
\begin{aligned}
x_t=\sqrt{\overline{\alpha_t}}x_0+\sqrt{1-\overline{\alpha_t}}\epsilon+\delta\tanh[\sqrt{c}\lambda_o^ct/T_0]x_0,\\
\end{aligned}
\end{equation}
where $\epsilon$ is Gaussian noise and $\delta$ is the radial popularity coefficient that controls the diffusion strength of each node in hyperbolic space. 
$T_{0}$ is a constant to control the speed of control of radial growth rate.$\lambda _{x}^c=\frac{2}{1+c\parallel{\mathbf{x}}\parallel ^{2}}$

Then, we discuss the content only on a cluster tangent plane. 
The main reason why the general diffusion model does not perform well on the graph is the decline of the fast signal-to-noise ratio. 
Inspired by directional diffusion model~\cite{yang2023directional}, we designate the direction of the geodesic between each cluster's center point and the north pole $\mathbf{o}$ as the target diffusion direction while imposing constraints for forward diffusion processes.
Specifically, the angular similarity constraints for each node $i$ can be obtained by: 

\begin{equation}\label{eq:constrain}
\begin{aligned}
&\mathbf{z}= \mathrm{sgn}\left(\mathrm{logmap}_{\mathbf{o}}^{c}\left(\mathbf{h}_{\mu_i}\right)\right)*\epsilon, \\
&\epsilon\sim\mathcal{N}\left(0,I\right),\\
\end{aligned}
\end{equation}
where $z$ represents the angle constrained noise,$\epsilon$ is the Gaussian noise, $\mathbf{h_{\mu_i}}$ is  the clustering center corresponding to the $i$-th node.

Combining the radial and angular constraints, our geometric diffusion process can be described as:
\begin{equation}\label{diff}
\begin{aligned}
x_t=\sqrt{\overline{\alpha_t}}x_0+\sqrt{1-\overline{\alpha_t}}\mathbf{z}+\delta\tanh[\sqrt{c}\lambda_o^ct/T_0]x_0,\\
\end{aligned}
\end{equation}

\begin{theorem}\label{thm:normal}
Let $x_t$ indicate the node $x$ at the $t$-step in the forward diffusion process Eq~\eqref{diff}. As $t \to \infty$, the low-dimensional latent representation $\mathbf{x}_{t}$ of node x satisfies:
\begin{equation}\label{eq:sampling}
\begin{aligned}
\lim_{t\to\infty}\mathbf{x}_{t}\sim\mathcal{N}_{f}\left(\delta\mathbf{x}_{0},I\right).
\end{aligned}
\end{equation}
where $\mathcal{N}_{f}$ is an approximate folded normal distribution. 
More detail and proof can be referred to in the Appendix~\ref{app:folded}.
\end{theorem}

Figure~\ref{fig:diff} illustrates examples of the diffusion process with/without geometric constraints in hyperbolic space. 
We can observe that by adding isotropic noise to the hyperbolic latent diffusion process, the final diffusion result is completely random noise.
In contrast, the hyperbolic latent diffusion process with geometric constraints can significantly preserve the anisotropy of the graph. 
In other words, after the graph diffusion, the result still preserves the important inductive bias of the graph below rather than the completely random noise, which will directly affect the performance and generation quality of the denoising process

\noindent \textbf{Training and generation. } 
Then, we follow the standard denoising process~\cite{ho2020ddpm,yang2023directional} and train a denoising network to simulate the process of reverse diffusion.
We use a denoising network architecture of DDM based on UNET for training  to predict $\mathbf{x}_0$, as follows:
\begin{equation}\label{Eq:loss}
\begin{aligned}
\mathcal{L}_{\mathrm{HDM}}=\mathbb{E}\left \| f_{\theta }\left(X_{t},A,t\right)-X_{0} \right \| ^{2}.
\end{aligned}
\end{equation}
Note that the loss function of our geometric diffusion model remains consistent with DDPM~\cite{ho2020ddpm} based on Theorem~\ref{thm:normal}. 
The proof refers to the Appendix ~\ref{app:diffusion details}.

Regarding the generation, we propose an efficient sampling method based on theorem \ref{thm:k}.
Furthermore, we demonstrate that it is possible to sample at once in the same tangent space instead of sampling in different cluster center tangent spaces to improve efficiency.
As to the denoising process, we adopt a denoising process that can be used in generalized diffusion models\cite{yang2023directional}. 
Specifically, where a recovery operator and a noise addition operator are abstracted for use in various diffusion methods.
All the specifics regarding each stage of the diffusion process, along with the theoretical derivation, are documented in the Appendix~\ref{app:diffusion details}.

Similar to other hyperbolic learning model~\cite{Krioukov2010Hyperbolic,HGCN_ChamiYRL19,HNN:GaneaBH18}, we utilize the \textit{Fermi-Dirac decoder}~\cite{Krioukov2010Hyperbolic,NickelK17Poincare} to compute the connection probability. 
The diffusion and reverse processes are summarized in Algorithm~\ref{Alg:training} and Algorithm~\ref{Alg:sampling}. 

\begin{algorithm}[t]
    \caption{Training of \modelname} 
    \label{Alg:training}
    \begin{algorithmic}
        \STATE {\bfseries Input:} Graph $\mathcal{G}=\{\mathbf{X},\mathbf{A}\}$; Number of training epochs $E$; 
        \STATE {\bfseries Parameter:}  $\theta$ initialization;\\
        \STATE {\bfseries Output:}Predicted raw embedding $\hat{x_{\mathbb{H}}}$
        \STATE Encoding node to hyperbolic space  $x_{\mathbb{H}}$ $\gets $ Eq.~\eqref{Eq:encoding};\\
        \STATE Compute $k$-clusters by h-Kmeans;\\
        \STATE Project the embeddings onto each $\mathcal{T}_{\mathbf{o}_{i \in \{ |k| \}}}$\\
        \FOR{$e=1$ {\bfseries to} $E$}
            \STATE Get the embeddings $x_{\mathbb{H}^t}$ of $t$-steps  Eq.~\eqref{diff} ;\\
    
            \STATE  Predict the raw embeddings $\hat{x_{\mathbb{H}}}$ ;\\
            \STATE Compute the loss $\mathcal{L}=\mathcal{L}_{\mathrm{HDM}}$$\gets $ Eq.~\eqref{Eq:loss}; \\
            \STATE Update $\theta \gets \theta-\eta \nabla \theta$. 
        \ENDFOR
    \end{algorithmic}  
\end{algorithm}

\noindent\textbf{Complexity Analysis}
Let $G=\left ( X, E \right )$ be one of the graphs set $G^s$, where $X$ is the $n$-dimensional node eigenvector and  $E$ is the $m*m$-dimensional adjacency matrix of the graph.  $s$ is the number of graphs in the graph set $G^s$. 
\textbf{Time complexity}: The time complexity of hyperbolic graph encoding is $O((1(t)+k)md)$. 
For the forward diffusion process, the complexity is $O(md)$. 
The training of denoising networks is essentially the same as other diffusion models and does not require additional computing time as $O(md)*1(t)$. 
Overall, the total time complexity of the diffusion process is $O(1(t)*2md)+O((k+2)md)$ in one epoch.
\textbf{Space complexity} In our approach, since we embed the graphs in hyperbolic space, each graph is represented as a $m*d$-dimensional vector in the hyperbolic space, which means that our diffusion scale is $O(smd)$. For a more detailed complexity analysis please refer to Appendix~\ref{app:complexity}.





\section{Experiment}
\label{sec:Experiment}
In this section, we conduct comprehensive experiments to demonstrate the effectiveness and adaptability of \modelname~\footnote{The code is available at \url{https://github.com/RingBDStack/HypDiff}.} in various datasets and tasks. 
We first presented the experimental settings and then showcased the results. 

\begin{table*}[htbp]
\caption{Summary of node classification Micro-F1 and Macro-F1 scores (\%) based on the average of five runs on synthetic and real-world datasets. 
(Result: average score ± standard deviation (rank); \textbf{Bold}: best; \underline{Underline}: runner-up.)}
\centering
\resizebox{\textwidth}{!}{
\begin{tabular}{c|cc|cc|cc|cc|cc|c}
\toprule
\multirow{3}{*}{\textbf{Method}} & \multicolumn{4}{c|}{\textbf{Synthetic Datasets}}    & \multicolumn{6}{c|}{\textbf{Real-world Datasets}}    & \multirow{3}{*}{\textbf{Avg. R.}}\\ 

\cline{2-11} 
   &  \multicolumn{2}{c}{SBM} &  \multicolumn{2}{c|}{BA} &  \multicolumn{2}{c}{Cora} &  \multicolumn{2}{c}{Citeseer} &  \multicolumn{2}{c|}{Polblogs} \\
   \cline{2-11} 
   &  Mi-F1 &  Ma-F1  &  Mi-F1 &  Ma-F1  &  Mi-F1 &  Ma-F1  &  Mi-F1 &  Ma-F1  &  Mi-F1 &  Ma-F1 \\
\midrule
VGAE   & 20.5±2.1 & 15.4±1.1 & 37.4±1.7 & 15.9±2.3 & 79.7±0.4 & 78.1±0.2 & 63.8±1.4 & 55.5±1.3 & 79.4±0.8 & 79.4±0.8 & 4.6  \\ 
ANE    & 39.9±1.1   & 33.9±1.8   & 46.0±3.0   & 19.3±2.7  & 69.3±0.1   & 66.4±0.1  & 50.2±0.1   & 49.5±0.6  & 80.8±0.1   &80.7±0.1  & 4.3  \\ 
GraphGAN   & 38.6±0.5   & 38.9±0.3   & 43.6±0.6   & 24.6±0.5  & 71.7±0.1   & 69.8±0.1  & 49.8±1.0   & 45.7±0.1  & 77.5±0.6   & 76.9±0.4 & 4.8  \\ 
\midrule
$\mathcal{P}$-VAE  & \underline{57.9±1.3}   & \underline{53.0±1.5}   & 38.4±1.4   & 20.0±0.3  & 79.6±2.2   & 77.5±2.5  & \textbf{67.9±1.7}   & \underline{60.2±1.9}  & 79.4±0.1   & 79.4±0.1   & 3.2  \\ 
Hype-ANE & 18.8±0.3   & 11.9±0.1   & \underline{56.9±2.4}   & \underline{31.6±1.2}  & \underline{80.7±0.1}   & \underline{79.2±0.3}  & 64.4±0.3   & 58.7±0.0  & \underline{83.6±0.4}   & \underline{83.6±0.4} & \underline{3.0}
  \\ 
\midrule
\textbf{\modelname~}    & \textbf{70.5±0.1}   & \textbf{69.4±0.1}   & \textbf{58.3±0.1}   & \textbf{40.0±0.1}  & \textbf{82.4±0.1}  & \textbf{81.2±0.1}  & \underline{67.8±0.2}   & \textbf{60.4±0.3}  & \textbf{85.7±0.1}   & \textbf{85.4±0.1}  & \textbf{1.1}\\ 
\bottomrule
\end{tabular}}
\label{table:nc_task}
\end{table*}

\begin{table*}[htbp]
\caption{Generation results about the MMD distance between the original and generated graphs.\\
(Result: scores (rank) and average rank;\textbf{Bold}: best; \underline{Underline}: runner-up.)}
\centering
\resizebox{\textwidth}{!}{
\begin{tabular}{c|ccc|ccc|ccc|ccc}
\toprule
\multirow{3}{*}{\textbf{Method}} & \multicolumn{6}{c|}{\textbf{Synthetic Datasets}}    & \multicolumn{6}{c}{\textbf{Real-world Datasets}}  \\ 
\cline{2-13} 

& \multicolumn{3}{c}{\textbf{Community}} & \multicolumn{3}{c|}{\textbf{BA-G}}    & \multicolumn{3}{c}{\textbf{MUTAG}}    & \multicolumn{3}{c}{\textbf{PROTRINS}}  \\ 
\cline{2-13} 
                        & Degree                & Cluster   & Spectre             & Degree    & Cluster            & Spectre   & Degree        & Cluster        & Spectre   & Degree        & Cluster        & Spectre  \\ 
\midrule
VGAE    & 0.365 & 0.025 &  0.507       & 0.775   & 1.214  & 0.398       & 0.255   & 2.000     & 0.744  & 0.705 & 0.979 & 0.700  \\ 
GraphRNN   & \textbf{0.002}  & 0.027 & \textbf{0.004}        & \textbf{0.122}    & 0.262   & 0.007       & 0.537   & \underline{0.013}   & 0.476  & \textbf{0.009}  & 0.071 & 0.017\\ 
\midrule
GDSS    & 0.094 & 0.031 & 0.052        & 0.978   & 0.468  & 0.917        & 0.074   & 0.021 & \textbf{0.003}    & 1.463  & 0.168 & \underline{0.013} \\ 
DiGress   & 0.226  & 0.158 & 0.194        & 0.654  & 1.171  &  0.268       & 0.100   & 0.351    & 0.082   & 0.108 & \underline{0.062} & 0.079\\ 
GraphGDP   &0.046 &0.016 & 0.042        & 0.698   & 0.188  &  \underline{0.053}       & 0.127   & 0.057    & 0.050   & 0.103 & 0.240 & 0.088\\ 
EDGE   & \underline{0.021} &\underline{0.013} & 0.040         & 0.282   &\textbf{ 0.010}  &  0.090       & \textbf{0.024}   & 0.597    & 0.468   & \underline{0.033} & 0.523 & 0.024\\ 
\midrule
\textbf{\modelname}    & \textbf{0.002} & \textbf{0.010} & \underline{0.028}        &\underline{ 0.216}   & \underline{0.021}
  & \textbf{0.004}        &\underline{0.048}   & \textbf{0.001} & \underline{0.040}  & 0.133  & \textbf{0.004} & \textbf{0.012}\\ 
\bottomrule
\end{tabular}}
\label{table:synthetic_mmd}
\end{table*}


\subsection{Datasets}
We estimate the capabilities of~\modelname~in various downstream tasks while conducting experiments on synthetic and real-world datasets. 
In addition, we construct and apply node-level and graph-level datasets for node classification and graph generation tasks. 
Statistics of the real-world datasets Table~\ref{tab:real} can be found in Appendix~\ref{app:datasets}. 
We elaborate on more details as follows. 

\textbf{Synthetic Datasets.}
We first use two well-accepted graph theoretical models, \textit{Stochastic Block Model (SBM)} and \textit{Barabási-Albert (BA)}, to generate a node-level synthetic dataset with 1000 nodes for node classification, respectively.
(1) \textbf{SBM} portrays five equally partitioned communities with the edge creation of intra-community $p=0.21$ and inter-community $q=0.025$ probabilities. 
(2) \textbf{BA} is grown by attaching new nodes each with random edges between 1 and 10.
Then we employ four generic datasets with different scales of nodes $\left | V \right |$ for graph generation tasks. 
Then, four datasets are generated for the graph-level task. 
(3) \textbf{Community} contains 500 two-community small graphs with $12\le \left | V \right | \le20$. 
Each graph is generated by the Erd\H{o}s-R{\'e}nyi model with the probability for edge creation $p=0.3$ and added $0.05\left | V \right | $ inter-community edges with uniform probability. 
(4) \textbf{Ego} comprises 1050 3-hop ego-networks extracted from the PubMed network with $\left | V \right | \le20$. 
Nodes indicate documents and edges represent their citation relationship.
(5) \textbf{Barabási-Albert (G)} is a generated graph-level dataset by the Barabási-Albert model (aka. BA-G to distinct node-level BA) with 500 graphs where the degree of each node is greater than four. 
(6) \textbf{Grid} describes 100 standard 2D grid graphs which have each node connected to its four nearest neighbors. 

\textbf{Real-world Datasets.}
We also carry out our experiments on several real-world datasets. 
For the node classification task, we utilize 
(1) two citation networks of academic papers including \textbf{Cora} and \textbf{Citeseer}, where nodes express documents and edges represent citation links, and 
(2) \textbf{Polblogs} dataset which is political blogs and is a larger size dataset we used.
With the graph generation task, we exploit four datasets from different fields. 
(3) \textbf{MUTAG} is a molecular network whose each graph denotes a nitro compound molecule. 
(4) \textbf{IMDB-B} is a social network, symbolizing the co-starring of the actors. 
(5) \textbf{PROTEINS} is a protein network in which nodes represent the amino acids and two nodes are connected by an edge if they are less than 6 Angstroms apart. 
(6) \textbf{COLLAB } is a scientific collaboration dataset, reflecting the collaboration of the scientists. 


\subsection{Experimental Setup}
\textbf{Baselines.} 
To evaluate the proposed \modelname~, we compare it with well-known or state-of-the-art graph learning methods which include: 
(1) \textbf{Euclidean graph representation methods}: 
VGAE~\cite{kipf2016variational} designs a variational autoencoder for graph representation learning.  
ANE~\cite{dai2018adversarial} trains a discriminator to align the embedding distribution with a predetermined fixed prior.  
GraphGAN~\cite{wang2018graphgan} learns the sampling distribution for negative node sampling from the graph. 
(2) \textbf{Hyperbolic graph representation learning}: 
$\mathcal{P}$-VAE~\cite{mathieu2019poincare} is a variational autoencoder utilizing the Poincaré ball model within hyperbolic geometric space.
Hype-ANE~\cite{HyperANE} is a hyperbolic adversarial network embedding model that extends ANE into hyperbolic geometric space.
(3) \textbf{Deep graph generative models}: 
VGAE~\cite{kipf2016variational} can be used for graph generation tasks by treating each graph as a batch size. 
GraphRNN~\cite{GraphRNN2018} is a deep auto-regressive generative model that focuses on graph representations under different node orderings. 
(4) \textbf{Graph diffusion generative models}: 
GDSS~\cite{jo2022GDSS} simultaneously diffuses node features and adjacency matrices to learn their scoring functions within the neural network correspondingly. 
DiGress~\cite{vignac2022digress} is a discrete denoising diffusion model that progressively recovers graph properties by manipulating edges.
GraphGDP~\cite{graphGDP} is a position-enhanced graph score-based diffusion model for graph generation.
EDGE~\cite{EDGE} is a discrete diffusion process for large graph generation.

\textbf{Settings.} \label{sec:config}
A fair parameter setting for the baselines is the default value in the original papers and for the training on new datasets make appropriate adjustments. 
For~\modelname, the encoder is 2-layer HGCN with $256$ representation dimensions, the edge dropping probability to 2\%, the learning rate to $0.001$, and hyperbolic curvature $c=1$. 
Additionally, the diffusion processing set diffusion strength $\delta$ as 0.5, and the number of 6 latent layers in denoising is 64, 128, 256, 128, 256, 128. 
We use Adam as an optimizer and set L2 regularization strength as 1e-5.
For the metric, we use the F1 scores of the node classification task and the maximum mean discrepancy scores of Degree, Cluster, and Spectre and the F1 score of precision-recall and density-coverage (F1 pr and F1 dc) to evaluate graph generation results. 

The richer experimental results under the other indicators are shown in Appendix~\ref{app:exper results}. 
All experiments adopt the implementations from the PyTorch Geometric Library and Deep Graph Library. 
The reported results are the average scores and standard deviations over $5$ runs. 
All models were trained and tested on a single Nvidia A100 40GB GPU. 

\subsection{Performance Evaluation}\label{sec:performance}
We show the F1 scores of the node classification task in Table~\ref{table:nc_task} and the statistics of MMD distance and F1 scores between the original and generated graph in the graph generation task in Table~\ref{table:synthetic_mmd} and Table~\ref{table:f1_prdc}.
A higher score reported in F1 indicates a more accurate prediction of the node and fidelity of the generated graph. At the same time, a smaller MMD distance suggests better generative capabilities of the model from the perspective of graph topological properties. 

\noindent\textbf{Node classification.} 
~\modelname~demonstrates superior performance which outperforms nearly all baseline models, achieving the highest ranking and revealing excellent generalization.  
This implies that~\modelname~can preserve essential properties within complex structures, enabling better distinctive and utility of the dependencies between nodes across hierarchical levels in hyperbolic space.

\noindent\textbf{Graph Generation.} 
Successively, we focused on validating the graph generation capability of~\modelname.
Using the finer-grained metrics, we consistently observed our approach's outstanding performance. 
More results are shown in Table~\ref{table:synthetic_mmd2}. 
We are further concerned with the fidelity and diversity of the generated results which yielded conclusions consistent with the previous and are reported in Table~\ref{table:f1_prdc}. 
Specifically, ~\modelname~depicts superior overall performance compared to the state-of-the-art model auto-regressive model GraphRNN and discrete diffusion method DiGress. 
Furthermore, our model can effectively capture the local structure through similarity constraints and achieve competitive performance on highly connected graph data (Community).

\subsection{Analysis of \modelname}
In this subsection, we present the experimental results to intuitively convey our discovery and initiate a series of discussions and analyses. 



\begin{figure*}[!t]
    \begin{minipage}{0.34\textwidth}
        \centering
        \includegraphics[height=4.3cm]{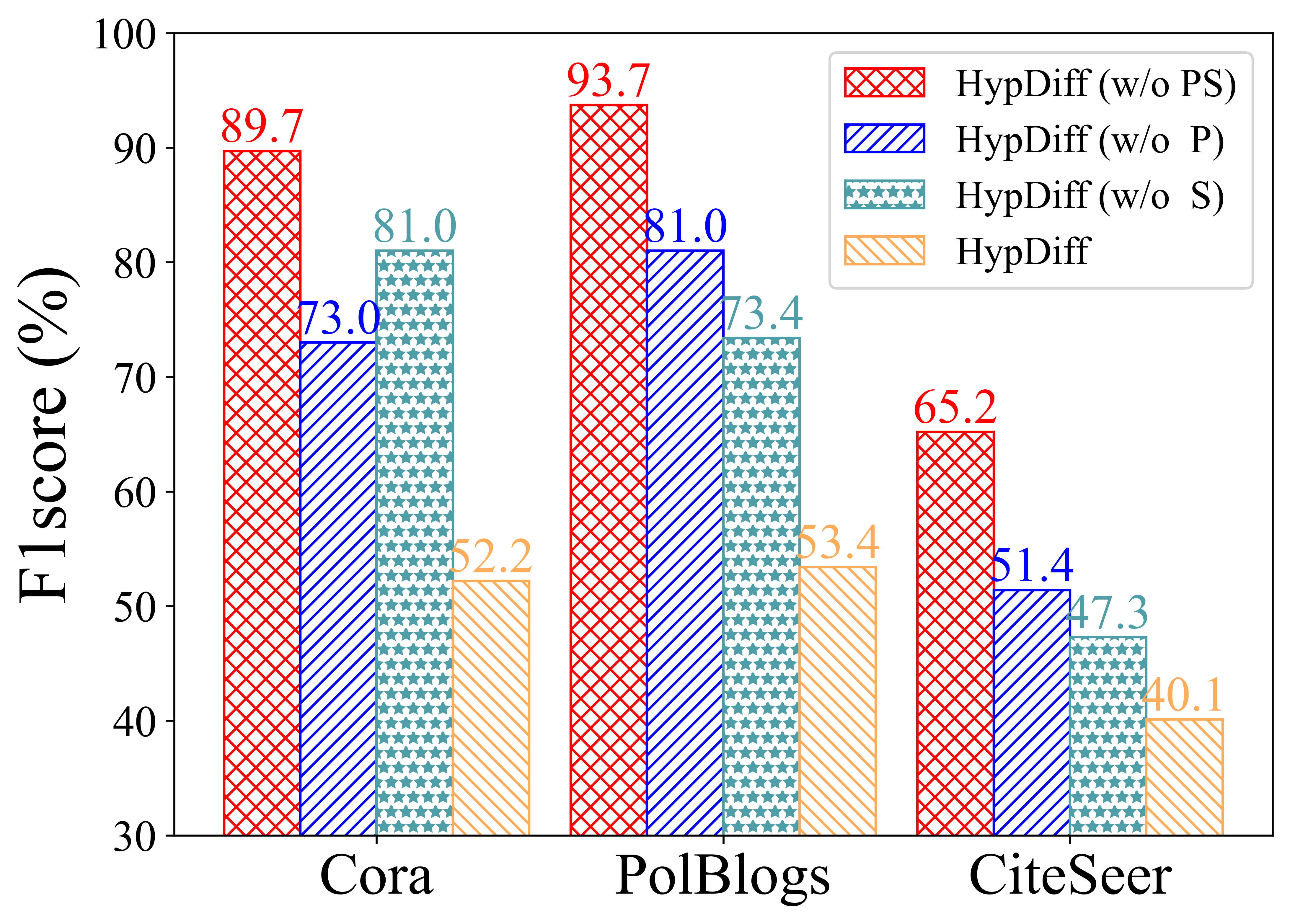}
        \vspace{-0.5cm}
        \caption{Ablation study results.}
        \label{fig:ablation}
    \end{minipage}
    \hspace{0.2cm}
    \begin{minipage}{0.65\textwidth}
        \centering
        \includegraphics[height=4.6cm]{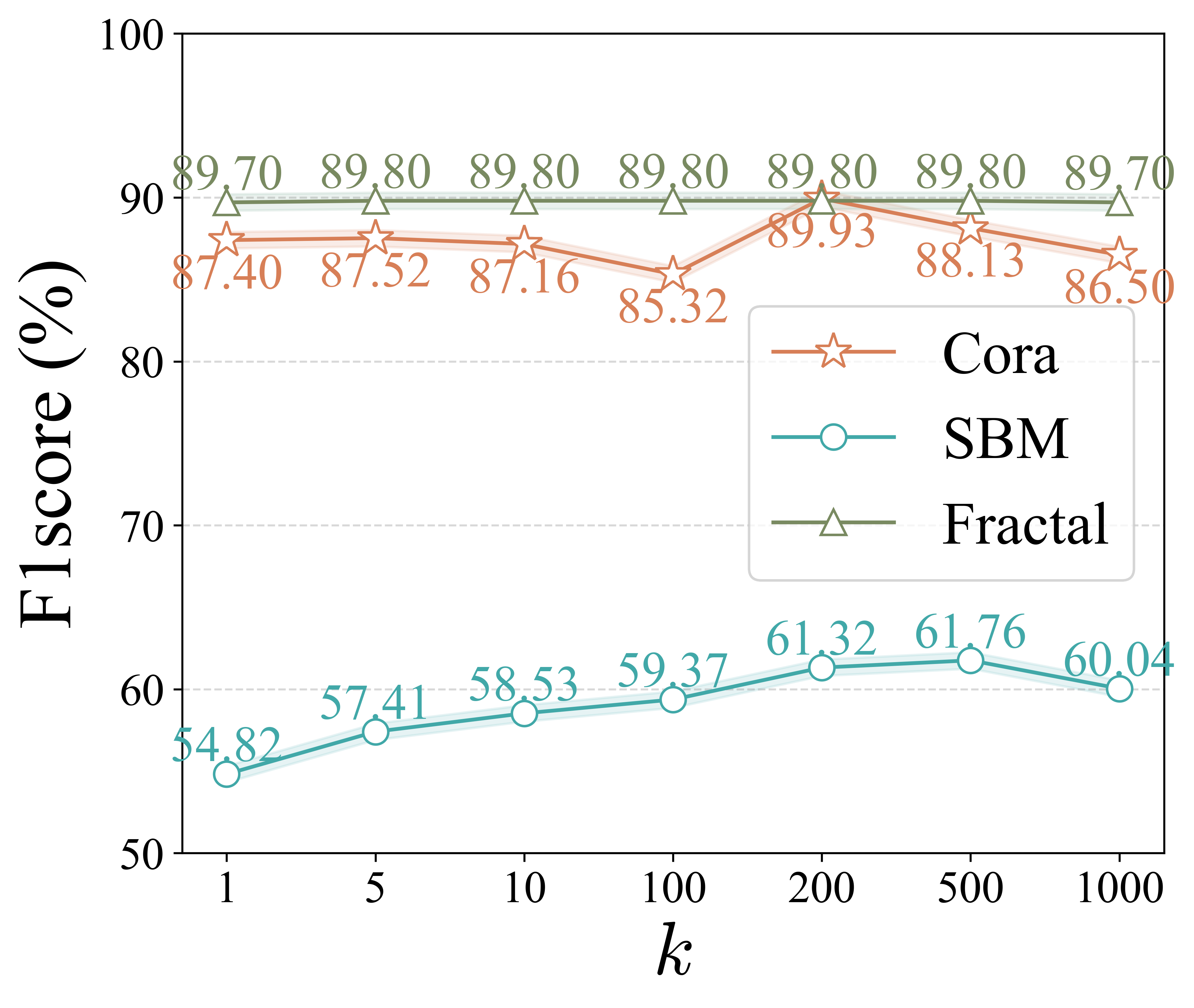}
        \includegraphics[height=4.6cm]{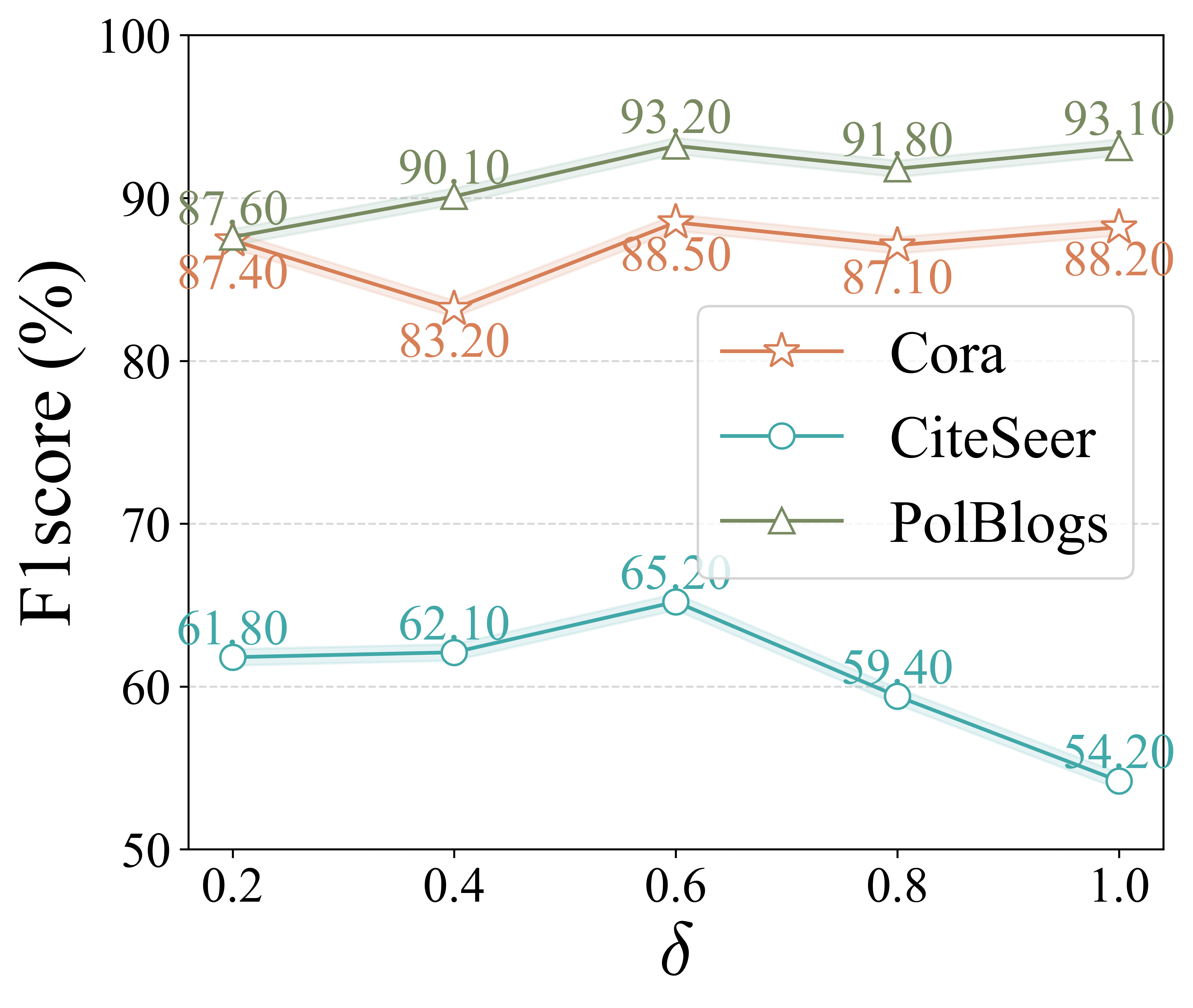}
        \vspace{-0.4cm}
        \caption{Sensitivity analysis of geometric constraints.}
        \label{fig:sensitivity}
    \end{minipage}
\end{figure*}

\begin{figure}[!t]\label{fig:efficiency}
\centering
\includegraphics[width=0.9\linewidth]{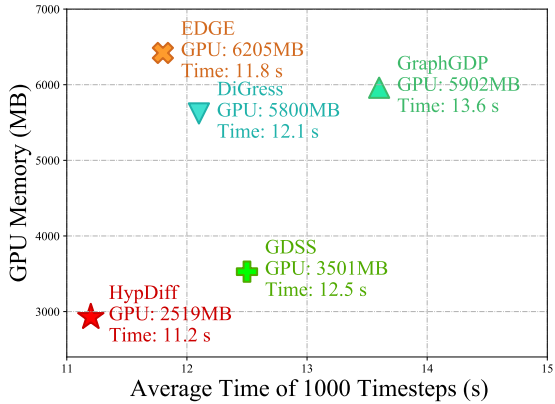}
\vspace{-0.25cm}
\centering
\caption{Efficiency analysis on IMDB-B for graph generation.}
\label{fig:efficiency}
\end{figure}
\noindent\textbf{Ablation Study.}
This study is to highlight the role of radial popularity diffusion and angular similarity diffusion constraints of \modelname.
We conducted experiments on three real-world datasets to validate the node classification performance and removed radial popularity (\textbf{HypDiff (w/o P)}), angular similarity (\textbf{HypDiff (w/o S)}) and total geometric prior(\textbf{HypDiff (w/o PS)}) components as the variant models. 
We show the results in Figure~\ref{fig:ablation}.
The radial popularity is evident in facilitating hyperbolic diffusion processes, thereby showcasing the advantage of hyperbolic geometry in capturing the underlying graph topology.
Furthermore, the angular similarity also significantly preserves the local structure of the graph, compensating for the limitations of hyperbolic space in capturing local connectivity patterns.
In summary, the hyperbolic geometric prior plays a crucial role in capturing non-Euclidean structures. 

\noindent\textbf{Sensitivity Analysis of Geometric Constraints.}
To investigate the impact of both the number of clusters $k$ and the geometric prior coefficient $\delta$ on the model performance, we conducted the sensitivity analysis on the real-world and synthetic graph datasets, respectively.
The number of clusters $k$ can be understood as the strength of the angular constraint, the results of three datasets with different structures are shown in Fig~\ref{fig:sensitivity} (Left).
Specifically, \textbf{Cora} has a real-world connected structure, \textbf{SBM} has a complex community structure, and \textbf{Fractal} has self-similarity and hierarchy properties.
It can be observed that $k$ has different sensitivities in different structured datasets, indicating that different graph structures have different approximate accuracies for anisotropy capture.
Correspondingly, the geometric prior coefficient $\delta$ can be understood as the strength of the radial constraint, the results of three real-world datasets are shown in Fig~\ref{fig:sensitivity} (Right).
The stronger the constraint, the smaller the diffusion step in the radial direction of the hyperbolic space.
It can be observed that the data set with a tree-like structure requires lower radial constraints, while the graph with high connectivity requires stronger radial constraints.
For the experimental setup and a more detailed analysis of the results please refer to Appendix~\ref{app:senstivity}.

\noindent\textbf{Diffusion Efficiency Analysis.}
We report the training time for our \modelname~and other graph diffusion baselines with the same configurations on IMDB-B. We conduct experiments with the hardware and software configurations listed in Section~\ref{sec:config}. We comprehensively report the results from the time and space costs of the diffusion process.
The result is shown in Figure~\ref{fig:efficiency}, our \modelname~ comprehensively outperforms other baselines in diffusion time and GPU memory cost. 
Compared with the discrete graph diffusion model, our model directly diffuses each node of the graph with structure-preserving based on the latent diffusion model, so the space complexity is much lower than that of the direct diffusion of discrete and sparse structural information(e.g. adjacent/Laplace matrix).
The performance of each dataset is in the Appendix \ref{gpu},

\noindent\textbf{Visualization.}
We compare the contributions of two diffusion generation models,~\modelname~and GDSS, to graph generation tasks by visualizing networks generated by five well-accepted graph theoretical models. 
We discuss and show the visualization as Figure~\ref{fig:visualization} in the Appendix~\ref{app:visualization}.

\section{Conclusion}
In this paper, we introduce hyperbolic geometric before solving the conflict problem between discrete graph data and continuous diffusion model, and propose a novel hyperbolic geometric diffusion model named~\modelname.
We propose an improved hyperbolic Gaussian noise generation method based on radial popularity to deal with the additive failure of Gaussian distributions in hyperbolic space.
The geometric constraints of angular similarity are applied to the anisotropic diffusion process, to preserve as much various local structure information as possible.
Extensive experiments conducted on both synthetic and real-world graphs demonstrate the comprehensive capability of \modelname.

\newpage
\section{Acknowledgments}
The corresponding authors are Jianxin Li and Xianxian Li. This paper is supported by the National Science and Technology Major Project of China (No.2022ZD0117800), and the National Natural Science Foundation of China (No.U21A20474 and 62302023). We owe sincere thanks to all co-authors for their valuable efforts and contributions. 

\section{Impact Statements}
This paper presents work whose goal is to advance the field of Machine Learning. There are many potential societal consequences of our work, none of which we feel must be specifically highlighted here.

\bibliography{reference}
\bibliographystyle{icml2024}

\newpage
\appendix
\onecolumn
\setcounter{table}{0}
\setcounter{footnote}{0}
\setcounter{figure}{0}
\setcounter{equation}{0}
\renewcommand{\thetable}{\ref*{app:proof}.\arabic{table}}
\renewcommand{\thefigure}{\ref*{app:proof}.\arabic{figure}}
\renewcommand{\theequation}{\ref*{app:proof}.\arabic{equation}}

\section{Summary of Notations.}\label{app:notation}

\begin{table}[h]
\caption{Summary of notations.}
\small
\centering
\begin{tabular}{c|l}
\toprule
\textbf{Symbol}    & \textbf{Description} \\ 
\midrule 
\textbf{$G$}     & Graph  \\ 
\textbf{$X$}     & Feature matrix  \\
\textbf{$A$}     & Adjacent matrix  \\  
\textbf{$d$}     & Dimension of the latent space  \\ 
\textbf{$n$}     & Number of nodes  \\ 
\textbf{$\mathbb{H}$}     & Hyperbolic space  \\ 
\textbf{$\mathbb{K}$}     & Klein model  \\ 
\textbf{$\dot{T}_{\mu}\mathbb{H}^{n}$}    & tangent space of node $\mu$ \\
\textbf{$c$}     & Hyperbolic curvature  \\
\textbf{$\eta $}     & Random variable  \\
\textbf{$\mathrm{expmap}_{\mu}^{c}(u)$}    & Exponential map. \\
\textbf{$\mathrm{logmap}_{\mathbf{\mu}}^{c}\left(x\right)$}     & Logarithmic map  \\
\textbf{$\lambda_{x}^{c}$}     & Curvature metric parameters \\

\textbf{$\oplus_{c}$}     & Möbius' addition  \\
\textbf{$\mathcal{N}_{\mathrm{H}}^{c}$}     & Wrapped normal distribution  \\
\textbf{$\mathcal{N}\left(0,I\right) $}     &  Gaussian distribution\\
\textbf{$\mathcal{N}_{f}\left(\delta\mathbf{x}_{0},I\right).$}     & Approximate folded normal distribution  \\
\textbf{$\sqrt{\overline{\alpha_{t}}} $}     & Diffusion coefficient  \\
\textbf{$\delta $}     & Radial constraint coefficient   \\
\textbf{$k$}     & Hyperbolic clustering parameter \\
\textbf{$x_0$}     & Initial diffusion embedding coordinates  \\
\textbf{$\epsilon $}     & Gaussian noise \\
\textbf{$z $}     & Angular constrained Gaussian noise \\
\textbf{$h $}     & Hyperbolic embedding coordinates \\

\bottomrule
\end{tabular}
\label{notion_description}
\end{table}

\section{Anisotropic Diffusion with/without Angular Noise}\label{app:snr}

We refer to the anisotropic angular noise mentioned in Directional Diffusion Model(DDM)~\cite{yang2023directional}: 
\begin{equation}
\begin{aligned}
\epsilon^{\prime}=\operatorname{sgn}(x_{0})\odot|\bar{\epsilon}|
\end{aligned}
\end{equation}
where $\epsilon$ denotes the Gaussian noise, $x_0$ denotes the graph nodes represent embedding coordinates.

\begin{figure}[h]
\centering
\includegraphics[width=0.5\linewidth]{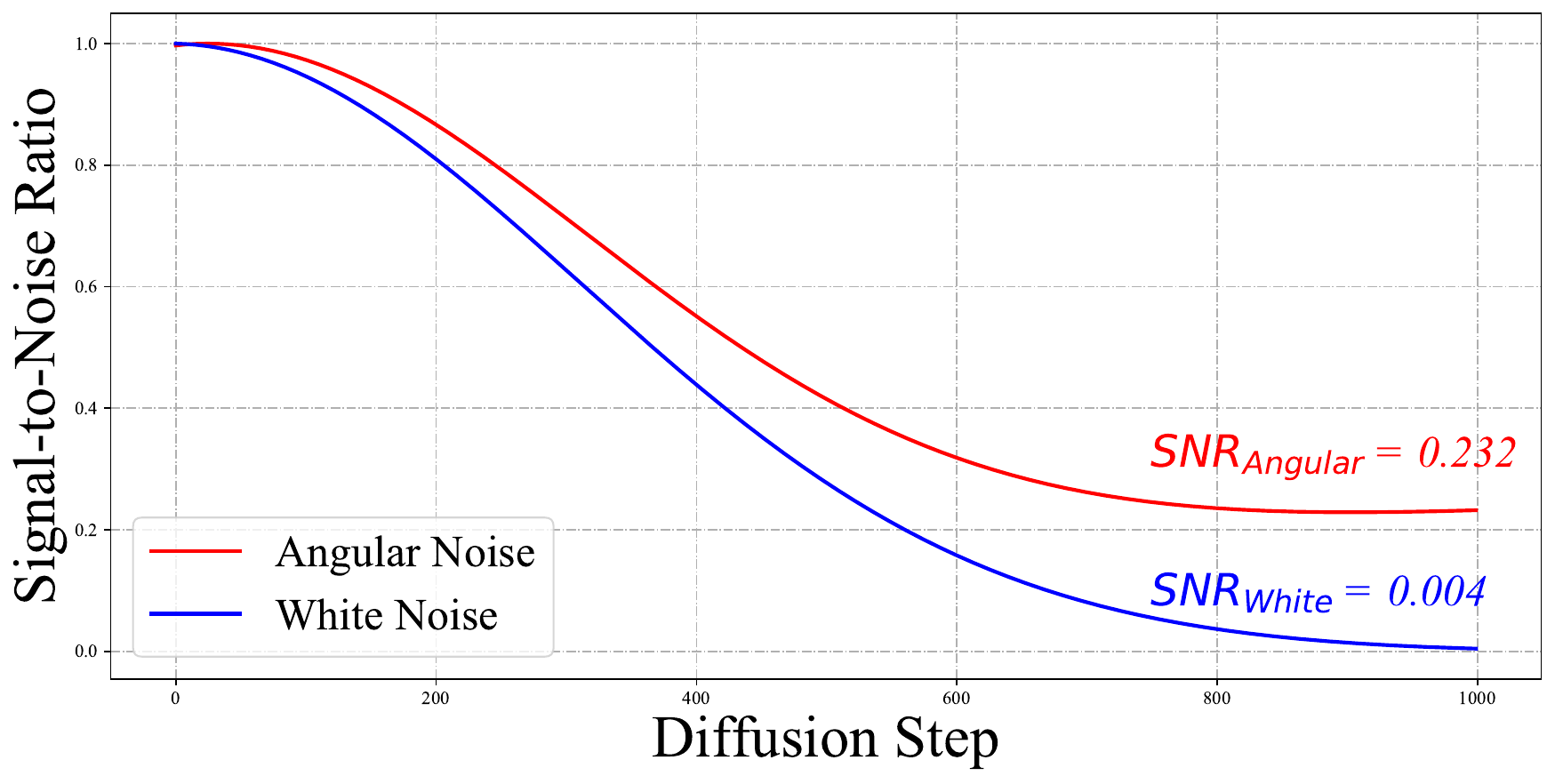}
\centering
\caption{Curve of Signal-to-Noise Ratio of different diffusion steps.}
\label{fig:snr2}
\end{figure}

Figure~\ref{fig:snr2} shows the SNR results of angular noise and white noise in the diffusion process, respectively. 
With the iterations (diffusion steps) increasing, the $SNR_{Angular}$(red) curve of angular noise is earlier than the $SNR_{White}$(blue) curve of white noise. 
The blue curve shows that the isotropic white noise rapidly masks the underlying anisotropic structure or signal during the standard diffusion process, which indicates the local details of the graph structure are greatly lost. 
\modelname~(red curve) benefits from the angular similarity constraint and can effectively preserve the anisotropic community structure during the diffusion process.

\section{Proof and Analysis}
\label{app:proof}
\subsection{Proof of the absence of additivity of the package normal distribution}\label{pro:add}
\noindent \textbf{Description of symbols.}
$\mathbb{B}_{c}^{d}$ is a d-dimentional Poincar\'e Ball space with curvature $c$.
$\mathbf{R}^d$ is a d-dimensional Euclidean space.$\mathbf{G(z)}$ is the metric tensor of the hyperbolic space.
$\mathbf{d_{p}^{c}}$ is the hyperbolic distance.
We first introduce the expression of metric tensor $\mathbf{G(z)}$ in hyperbolic space and the hyperbolic distance $\mathbf{d_{p}^{c}}$:
\begin{equation}
    \begin{aligned}
        &\mathbf{G(z)}=\begin{pmatrix}1&0\\0&\left(\frac{\sinh(\sqrt{c}r)}{\sqrt{c}r}\right)^2I_{d-1}\end{pmatrix},\\
        &\mathbf{d_{p}^{c}}(\boldsymbol{z},\boldsymbol{y})=\frac{1}{\sqrt{c}}\cosh^{-1}\left(1+2c\frac{\|z-\boldsymbol{y}\|^{2}}{(1-c\left\|\boldsymbol{z}\right\|^{2})(1-c\left\|\boldsymbol{y}\right\|^{2})}\right).
    \end{aligned}
\end{equation}
Following the $\mathcal{P}$-VAE~\cite{NickelK17Poincare} we can then obtain the differential and integral operators by transforming under hyperbolic polar coordinates and Euclidean space as:
\begin{equation}
\begin{aligned}
ds_{\mathbb{B}_{c}^{d}}^{2}& =(\lambda_{z}^{c})^{2}(dz_{1}^{2}+\cdots+dz_{d}^{2})=\frac{4}{\left(1-c\|x\|^{2}\right)^{2}}dz^{2}  \\
&=\frac{4}{(1-c\rho^{2})^{2}}(d\rho^{2}+\rho^{2}ds_{\mathrm{S}^{d-1}}^{2}),
\end{aligned}
\end{equation}
let $\mathbf{r}=\mathbf{d_{p}^{c}}$, we have
\begin{equation}
\begin{aligned}
\mathbf{r}&=\int_{0}^{\rho}\lambda_{t}^{c}dt =\int_{0}^{\rho}\frac{2}{1-ct^{2}}dt =\int_{0}^{\sqrt{c}\rho}\frac{2}{1-t^{2}}\frac{dt}{\sqrt{c}}\\
&=\frac{2}{\sqrt{c}}\tanh^{-1}(\sqrt{c}\rho).
\end{aligned}
\end{equation}
Then, we have
\begin{equation}
\begin{aligned}
ds_{\mathbb{B}_{c}^{d}}^{2}& =\frac{4}{(1-c\rho^{2})^{2}}\frac{1}{4}(1-c\rho^{2})^{2}dr^{2}+\left(2\frac{\rho}{1-c\rho^{2}}\right)^{2}ds_{\mathrm{S}^{d-1}}^{2}  \\
&=dr^{2}+\left(2\frac{\frac{1}{\sqrt{c}}\tanh(\sqrt{c}\frac{r}{2})}{1-c\left(\frac{1}{\sqrt{c}}\tanh(\sqrt{c}\frac{r}{2}\right)^{2}}\right)^{2}ds_{\mathsf{S}^{d-1}}^{2} \\
&=dr^{2}+\left(\frac{1}{\sqrt{c}}\sinh(\sqrt{c}r)\right)^{2}ds_{\mathsf{S}^{d-1}}^{2},
\end{aligned}
\end{equation}
when $\mathbf{c}\rightarrow\mathbf{0}$,it recovers the Euclidean line element as
\begin{equation}
\begin{aligned}
\mathbf{ds_{\mathbf{R}^d}^2}=\mathbf{dr^2+r^2ds_{\mathbf{S}^{d-1}}^2}.
\end{aligned}
\end{equation}
Then, we have the integral calculation as
\begin{equation}
\begin{aligned}
\int_{\mathbb{B}_{c}^{d}}f(\boldsymbol{z})d\mathcal{M}(\boldsymbol{z})& =\int_{\mathbb{B}_{c}^{d}}f(z)\sqrt{|G(z)|}dz  \\
&=\int_{{\mathcal T}_{\mu}{\mathcal B}_{c}^{d}\cong{\mathcal R}^{d}}f(\boldsymbol{v})\sqrt{|G(\boldsymbol{v})|}d\boldsymbol{v} \\
&=\int_{\mathbb{R}_{+}}\int_{\mathbb{S}^{d-1}}f(r)\sqrt{|G(r)|}drr^{d-1}ds_{\mathbf{S}d-1} \\
&=\int_{\mathbb{R}_{+}}\int_{\mathbb{S}^{d-1}}f(r)\left(\frac{\sinh(\sqrt{c}r)}{\sqrt{c}r}\right)^{d-1}drr^{d-1}ds_{\mathbf{S}^{d-1}} \\
&=\int_{\mathbb{R}_{+}}\int_{\mathbb{S}^{d-1}}f(r)\left(\frac{\sinh(\sqrt{c}r)}{\sqrt{c}}\right)^{d-1}drds_{\mathbb{S}^{d-1}}.
\end{aligned}
\end{equation}
\begin{proof}
We introduce the probability density distribution of the wrapped normal distribution.
The wrapped normal distribution is mapped to a hyperbolic space by taking the normal distribution in the tangent plane $\mathbf{\mathcal{T}_{\mu}\mathbb{B}_{c}^{d}}$ through the exponential map~\cite{said2014new}. One can obtain samples as follows:
\begin{equation}
\begin{aligned}
\mathbf{z}=\exp_{\mu}^{c}\left(G(\mu)^{-\frac{1}{2}}v\right)=\exp_{\mu}^{c}\left(\frac{v}{\lambda_{\mu}^{c}}\right),\mathrm{with}v\sim\mathcal{N}(\cdot|\mathbf{0},\Sigma).
\end{aligned}
\end{equation}

\noindent \textbf{Anisotropic.} 
In the anisotropic settings, its probability density can be given by:
\begin{equation}
\begin{aligned}
\mathcal{N}_{\mathbb{B}_{c}^{d}}^{\mathrm{W}}(z|\mu,\Sigma)& =\mathcal{N}\left(G(\boldsymbol{\mu})^{1/2}\log_{\boldsymbol{\mu}}(\boldsymbol{z})\left|\mathbf{0},\Sigma\right)\left(\frac{\sqrt{c}d_{p}^{c}(\boldsymbol{\mu},\boldsymbol{z})}{\sinh(\sqrt{c}d_{p}^{c}(\boldsymbol{\mu},\boldsymbol{z}))}\right)^{d\boldsymbol{-}1}\right.   \\
&=\mathcal{N}\left(\lambda_{\mu}^{c}\log_{\mu}(z)\big|\mathbf{0},\Sigma\right)\left(\frac{\sqrt{c}d_{p}^{c}(\boldsymbol{\mu},\boldsymbol{z})}{\sinh(\sqrt{c}d_{p}^{c}(\boldsymbol{\mu},\boldsymbol{z}))}\right)^{d\boldsymbol{-}1}.
\end{aligned} 
\end{equation}
We can plug its density with introducing the variable $v=r\alpha=\lambda_{\mu}^{c}\log_{\mu}(z)$  and utilizing the metric tensor, and we have
\begin{equation}
\begin{aligned}
&\int_{\mathbb{B}_{c}^{d}}\mathcal{N}_{\mathbb{B}_{c}^{d}}^{\mathbf{W}}(z|\mu,\Sigma)d\mathcal{M}(z)\\
=&\int_{\mathcal{T}_{\mu}\mathbb{B}_{c}^{d}\cong\mathbb{R}^{d}}\mathcal{N}\left(\boldsymbol{v}\mid\boldsymbol{0},\Sigma\right)\left(\frac{\sqrt{c}\left\|\boldsymbol{v}\right\|_{2}}{\sinh(\sqrt{c}\left\|\boldsymbol{v}\right\|_{2})}\right)^{d\boldsymbol{-}1}\sqrt{\left|G(\boldsymbol{v})\right|}d\boldsymbol{v}  \\
=&\int_{\mathbb{R}^{d}}\mathcal{N}\left(\boldsymbol{v}\mid\boldsymbol{0},\Sigma\right)\left(\frac{\sqrt{c}\left\|\boldsymbol{v}\right\|_{2}}{\sinh(\sqrt{c}\left\|\boldsymbol{v}\right\|_{2})}\right)^{d\boldsymbol{-}1}\left(\frac{\sinh(\sqrt{c}\left\|\boldsymbol{v}\right\|_{2})}{\sqrt{c}\left\|\boldsymbol{v}\right\|_{2}}\right)^{d\boldsymbol{-}1}d\boldsymbol{v} \\
=&\int_{\mathbb{R}^{d}}\mathcal{N}\left(\boldsymbol{v}\mid\mathbf{0},\Sigma\right)d\boldsymbol{v}.
\end{aligned}
\end{equation}

Next, we derive whether the sum of two independent wrapped normally distributed variables still satisfies the wrapped normal distribution.
\begin{equation}
\begin{aligned}
&\mathcal{N}_{\mathbb{B}_{c}^{d}}^{\mathbf{W}}(z_{1}|\mu_{1},\Sigma_{1})*\mathcal{N}_{\mathbb{B}_{c}^{d}}^{\mathbf{W}}(z_{2}|\mu_{2},\Sigma_{2})\\
=&\int_{\mathbb{B}_{c}^{d}}\mathcal{N}_{\mathbb{B}_{c}^{d}}^{\mathbf{W}}(z-z_{2}|\mu_{1},\Sigma_{1})\mathcal{N}_{\mathbb{B}_{c}^{d}}^{\mathbf{W}}(z_{2}|\mu_{2},\Sigma_{2})d\mathcal{M}(z_{2})\\
=&\int_{\mathbb{R}^{d}}\mathcal{N}\left(\boldsymbol{v-v_{2}}\mid\mathbf{0},\Sigma_{1}\right)\mathcal{N}\left(\boldsymbol{v_{2}}\mid\mathbf{0},\Sigma_{2}\right)\left(\frac{\sqrt{c}\left\|\boldsymbol{v-v_{2}}\right\|_{2}}{\sinh(\sqrt{c}\left\|\boldsymbol{v-v_{2}}\right\|_{2})}\right)^{d\boldsymbol{-}1}d\boldsymbol{v_{2}}\\
&\mathcal{N}_{\mathbb{B}_{c}^{d}}^{\mathbf{W}}(z_{1}|\mu_{1},\Sigma_{1})*\mathcal{N}_{\mathbb{B}_{c}^{d}}^{\mathbf{W}}(z_{2}|\mu_{2},\Sigma_{2})\nsim\mathcal{N}_{\mathbb{B}_{c}^{d}}^{\mathbf{W}}(z|\mu,\Sigma).
\end{aligned}
\end{equation}

\textbf{Isotropic.} In the isotropic setting, the density of the wrapped normal is given by:
\begin{equation}
\begin{aligned}
&\mathcal{N}_{\mathsf{B}_{\mathsf{c}}}^{\mathsf{W}}(z|\boldsymbol{\mu},\sigma^{2})=\frac{d\nu^{\mathsf{W}}(z|\boldsymbol{\mu},\sigma^{2})}{d\mathcal{M}(\boldsymbol{z})}\\
&=(2\pi\sigma^{2})^{-d/2}\exp\left(-\frac{d_{p}^{c}(\boldsymbol{\mu},\boldsymbol{z})^{2}}{2\sigma^{2}}\right)\left(\frac{\sqrt{c}d_{p}^{c}(\boldsymbol{\mu},\boldsymbol{z})}{\sinh(\sqrt{c}d_{p}^{c}(\boldsymbol{\mu},\boldsymbol{z}))}\right)^{d-1}.
\end{aligned}
\end{equation}
Its integral form can be given by:\\
\begin{equation}
\begin{aligned}
&\int_{_{B_{c}^{d}}}\mathcal{N}_{_{E_{c}^{d}}}^{\mathrm{W}}(z|\boldsymbol{\mu},\sigma^{2})d\mathcal{M}(z)\\
=&\int_{_{R_{+}}}\int_{_{S}^{d-1}}\frac{1}{Z^{\mathrm{R}}}e^{-\frac{r^{2}}{2\sigma^{2}}}r^{d-1}drds_{_{S}d-1},
\end{aligned}
\end{equation}
where $Z^R$ is the constant, and it is defined as
\begin{equation}
\begin{aligned}
Z^{\mathrm{R}}=&\zeta \binom{d-1}{k}e^{\frac{(d-1-2k)^2}{2}c\sigma^2}\left[1+\operatorname{erf}\left(\frac{(d-1-2k)\sqrt{c}\sigma}{\sqrt{2}}\right)\right],\\
\zeta =& \frac{2\pi^{d/2}}{\Gamma(d/2)}\sqrt{\frac{\pi}{2}}\sigma\frac{1}{(2\sqrt{c})^{d-1}}\sum_{k=0}^{d-1}(-1)^k.
\end{aligned}
\end{equation}

Thus, we can derive its additivity:
\begin{equation}
\begin{aligned}
&\mathcal{N}_{\mathbb{B}_{c}^{d}}^{\mathbf{W}}(z_{1}|\mu_{1},\Sigma_{1})*\mathcal{N}_{\mathbb{B}_{c}^{d}}^{\mathbf{W}}(z_{2}|\mu_{2},\Sigma_{2})\\
=&\int_{\mathbb{B}_{c}^{d}}\mathcal{N}_{\mathbb{B}_{c}^{d}}^{\mathbf{W}}(z-z_{2}|\mu_{1},\Sigma_{1})\mathcal{N}_{\mathbb{B}_{c}^{d}}^{\mathbf{W}}(z_{2}|\mu_{2},\Sigma_{2})d\mathcal{M}(z_{2})\\
=&\int_{_{R_{+}}}\int_{_{S}^{d-1}}\frac{1}{Z^{\mathrm{R^2}}}e^{-\frac{(r-r_{2})^{\mathrm{2} }}{2\sigma^{2}}}(r-r_{2})^{d-1} \gamma^c_p e^{-\frac{(r_{2})^{\mathrm{2} }}{2\sigma^{2}}}(r_{2})^{d-1}drds_{_{S}d-1}\\
=&\int_{_{R_{+}}}\int_{_{S}^{d-1}}\frac{1}{Z^{\mathrm{R^2}}}e^{-\frac{(r^{2}-2rr_{2}+2r_{2}^{2})}{2\sigma^{2}}}(r_{2}(r-r_{2}))^{d-1} \gamma^c_p drds_{_{S}d-1}\\
&\gamma^c_p=\left(\frac{\sqrt{c}d_{p}^{c}(\boldsymbol{\mu_{1}},\boldsymbol{z_{1}})}{\sinh(\sqrt{c}d_{p}^{c}(\boldsymbol{\mu_{1}},\boldsymbol{z_{1}}))}\right)^{d-1}\\
&\mathcal{N}_{\mathbb{B}_{c}^{d}}^{\mathbf{W}}(z_{1}|\mu_{1},\Sigma_{1})*\mathcal{N}_{\mathbb{B}_{c}^{d}}^{\mathbf{W}}(z_{2}|\mu_{2},\Sigma_{2})\nsim\mathcal{N}_{\mathbb{B}_{c}^{d}}^{\mathbf{W}}(z|\mu,\Sigma).
\end{aligned}
\end{equation}
Thus, we have demonstrated the lack of additivity in the wrapped normal distribution.
\end{proof}
\subsection{Proof of Theorem~\ref{thm:k}}\label{proof:k}
In hyperbolic geometry, four commonly used equivalent models are the Klein model, the Poincare disk model, the Lorentz model, and the Poincare half-plane model. For our analysis in this study, we utilize the Lorentz model.~\cite{ungar1999hyperbolic}

Lorentz model $\mathbb{H}^n$ can be denoted by a set of points $z(z\in\mathbb{R}^{n+1})$ through Lorentzian
product:
\begin{equation}
\begin{aligned}
\langle z,z'\rangle_{\mathcal L}=-z_{0}z_{0}^{\prime}+\sum_{i=1}^{n}z_{i}z_{i}^{\prime},
\end{aligned}
\end{equation}
\begin{equation}
\begin{aligned}
\mathbb{H}^n=\left\{z\in\mathbb{R}^{n+1}\colon\langle z,z\rangle_{\mathcal{L}}=\frac{1}{c},z_0>0\right\}.
\end{aligned}
\end{equation}

Tangent space 
$T_{\boldsymbol{\mu}}\mathbb{H}^n$ is the tangent space of $\mathbb{H}^n$ at
$\mu$. $T_{\boldsymbol{\mu}}\mathbb{H}^n$ can be represented as the set of points that satisfy the orthogonality relation with respect to the Lorentzian product:
\begin{equation}
\begin{aligned}
T_{\mu}\mathbb{H}^{n}:=\{\boldsymbol{u}:\langle\boldsymbol{u},\boldsymbol{\mu}\rangle_{\mathcal{L}}=0\}.
\end{aligned}
\end{equation}

\textbf{Parallel transport and inverse parallel transport.}
For an arbitrary pair of point $\mu,\nu\in{\mathbb{H}^{n}}$, the parallel transport from $\nu$ to $\mu$ is defined as a map $\mathrm{PT}_{\nu\rightarrow\mu}$ from $T_{\boldsymbol{\nu}}\mathbb{H}^n$  to $T_{\boldsymbol{\mu}}\mathbb{H}^n$ that carries a vector in  $T_{\boldsymbol{\nu}}\mathbb{H}^n$ along the geodesic from $\nu$ to $\mu$ without changing its metric tensor.

The explicit formula for the parallel transport on the Lorentz model is given by:
\begin{equation}
\begin{aligned}
\operatorname{PT}^c_{\nu\to\mu}(v)=v+\frac{\langle\mu-\alpha\nu,v\rangle_{\mathcal{L}}}{\alpha+1}(\nu+\mu),
\end{aligned}
\end{equation}
where $\alpha=-\langle\nu,\mu\rangle_{\mathcal{L}}$.The inverse parallel transport is given by:
\begin{equation}
\begin{aligned}
v=\mathrm{PT}^c_{\mu\rightarrow\nu}(u).
\end{aligned}
\end{equation}

\textbf{Exponential map and Logarithmic map.}
Exponential map $exp_mu$:$T_{\boldsymbol{\mu}}\mathbb{H}^n\to\dot{\mathbb{H}^n}$ is a map that we can use to project a vector $\nu$ in $T_{\boldsymbol{\mu}}\mathbb{H}^n$ to $\mathbb{H}^n$.The explicit formula for the exponential map on the Lorentz model is  given by:
\begin{equation}
\begin{aligned}
z=\exp^c_{\boldsymbol{\mu}}(\boldsymbol{u})=\cosh\left(\|\boldsymbol{u}\|_{\mathcal{L}}\right)\boldsymbol{\mu}+\sinh\left(\|\boldsymbol{u}\|_{\mathcal{L}}\right)\frac{\boldsymbol{u}}{\|\boldsymbol{u}\|_{\mathcal{L}}}.
\end{aligned}
\end{equation}
The logarithmic map is defined to compute the inverse of the exponential map, mapping the point back to the tangent space. It is given by:
\begin{equation}
\begin{aligned}
u=\log^c_{\boldsymbol{\mu}}(z)=\frac{\operatorname{arccosh}(\alpha)}{\sqrt{\alpha^{2}-1}}(z-\alpha\boldsymbol{\mu}),
\end{aligned}
\end{equation}
where $\alpha=-\langle\mu,z\rangle_{\mathcal{L}}$.

\textbf{Klein model.}
This model of hyperbolic space is a subset of $R^n$ given by $K^n$, and a point in the Klein model can be obtained from the corresponding point in the hyperbolic model by projection:
\begin{equation}
\begin{aligned}
\pi_{\mathbb{H}\to\mathbb{K}}(\mathbf{x})_i=\frac{x_i}{x_{0}}.
\end{aligned}
\end{equation}
with its inverse given by
\begin{equation}
\begin{aligned}
\pi_{\mathbb{K}\to\mathbb{H}}^{-1}(\mathbf{x})=\frac1{\sqrt{1-\left\|\mathbf{x}\right\|^2}}(1,\mathbf{x}).
\end{aligned}
\end{equation}

An interesting approach is that we can use the angle-preserving nature of the Klein model to construct a mapping from the Lorentz tangent plane to the Klein model via the spherical pole mapping $\mathrm{P}^c$:
\begin{equation}
\begin{aligned}
k_j=\mathrm{P}^c(x_j)=\frac1{-c+\sum_{i=1}^{n}x_{i}^2}(x_j^2).
\end{aligned}
\end{equation}
\begin{proof}
    In our model, $m$ represents the number of clusters and $h_i(h_i\in\mathbb{H}^n)$ represents the points in class $i(i\in 1,2,3,...,m)$,
 $\mu_i$ denotes the hyperbolic center of the cluster $i$.Each point $h_i$ will be mapped to the tangent plane $T_{\boldsymbol{\mu_i}}\mathbb{H}^n$ of its center $\mu_i$. 
 Let $x_i$ denote the point after mapping to the tangent plane, it can be calculated by:
 \begin{equation}
\begin{aligned}
 x_i=\log^{c}_{\boldsymbol{\mu_i}}(h_i)
 \end{aligned}
\end{equation}
If we consider all the points as being in the tangent plane to the North Pole $T_{\boldsymbol{o}}\mathbb{H}^n$, then their corresponding coordinates are:
\begin{equation}
\begin{aligned}
X=(x_1,x_2,...,x_m)
 \end{aligned}
\end{equation}
For a curvature $c_i$, if the following equation is satisfied:
\begin{equation}
\begin{aligned}
\log^{c_i}_{\boldsymbol{o}}(h_i)=\mathrm{PT}^{c_0}_{o\rightarrow \mu_i}(\log^{c_0}_{\boldsymbol{o}}(h_i))
\end{aligned}
\end{equation}
 then it is possible to transform the tangent planes from the various centers to the tangent plane at the North Pole and unify them into the Klein model. The mapping point $k$ is given by:
\begin{equation}
\begin{aligned}
k_i&=\mathrm{P}^{c_0}(\log^{c_0}_{\boldsymbol{o}}(h_i))\\
&=\mathrm{P}^{c_0}(\mathrm{PT}^{c_0}_{\mu_i\rightarrow o}(\log_{\boldsymbol{\mu_i}}(h_i)))\\
&=\mathrm{P}^{c_0}(\mathrm{PT}^{c_0}_{\mu_i\rightarrow o}(x_i))\\
&=\pi_{\mathbb{H}\to\mathbb{K}}(\mathbf{h_i})\\
&=\mathrm{P}^{c_i}(\log^{c_i}_o(h_i)).
\end{aligned}
\end{equation}
\end{proof}

This implies that our approach essentially involves mathematically projecting points to approximate a Klein model comprising multiple curvatures. The process represents a topological reconstruction of the geometric space derived from the original graph structure, thereby enhancing our ability to capture the geometric properties inherent in the original graph.

\section{Hyperbolic Geometric Priori of Graph}\label{app:hyperbolic}
The Popularity-Similarity Optimization (PSO) model ~\cite{papadopoulos2012popularity} is a generative network model to describe how random geometric graphs grow in hyperbolic Spaces to optimize the trade-off between node prevalence (abstracts by radial coordinates) and similarity (expressed by angular coordinate distances), which exhibits many common structural and dynamic features of realistic networks, such as clustering, small-worldness, scale-freeness and rich-clubness. 
It uses the node's birth time $t=1,2,\cdots, T$ to proxy popularity, giving preference to older nodes that are more likely to become popular and attract connections, which is similar to Key Opinion Leaders (KOLs) in social networks
The angular distances between nodes denote their similarity distances by using cosine similarity or any other measure. 
The node $t$ can be represented as polar coordinates $(r_t,\theta_t)$, and the objective is to establish new connections while optimizing the product between popularity and similarity. 
To simulate a realistic graph with scale-freeness (power-law degree distribution), let radial coordinate $r_t$ of the new node $t$ have $r_t = \ln t$, indicating that the node $t$ prefers to connect popular nodes.
In hyperbolic geometric space, the new node distance can represent a convenient single-metric by using a combination of the two geometric attractiveness priors, \textit{radial popularity} and \textit{angular similarity}. 
\begin{figure*}[h]
\centering
\subfigure[Wrapped Gaussian distributions in hyperbolic space.]{
\includegraphics[width=0.45\linewidth]{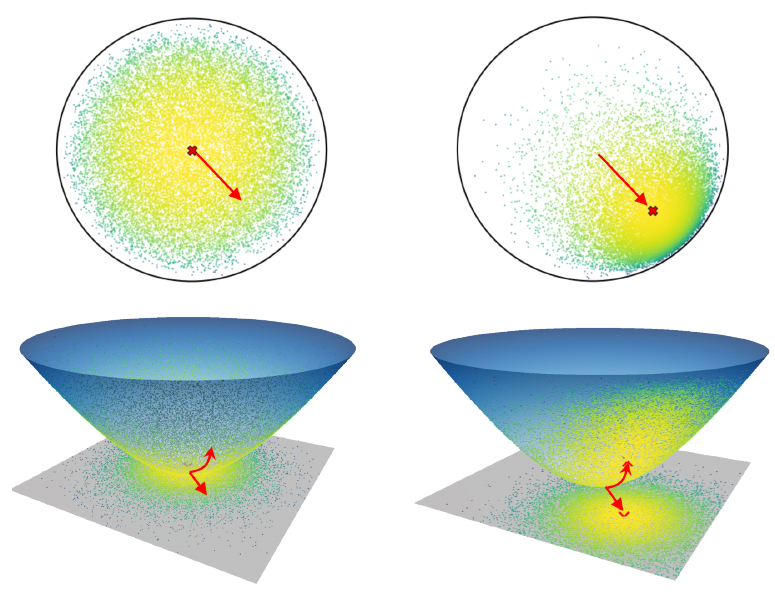}\label{fig:GEO-1}
}
\subfigure[Angular distributions in hyperbolic space.]{
\includegraphics[width=0.46\linewidth]{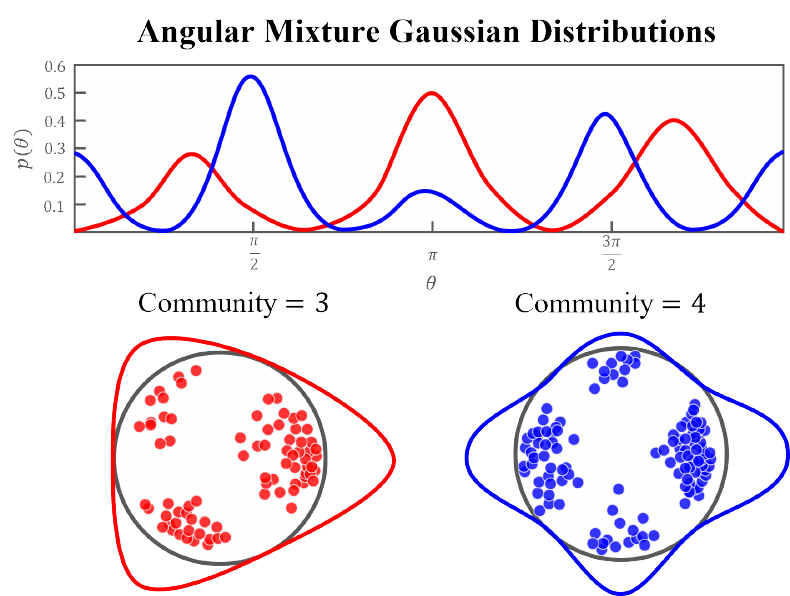}\label{fig:GEO-2}
}
\centering
\caption{(a) The Wrapped Gaussian distribution is commonly used to generate noise in hyperbolic space. 
As the $\mu$ approaches the edge of the Poincar\'e disk, the probability density at the edge of the disk increases, while the probability density at the center becomes sparse. It indicates a geometric interpretation for the disk radius, that there are always fewer nodes with popularity. 
(b) In hyperbolic space, the sectors of the disk are closely related to the community structure in the graph, and noise that follows a mixed Gaussian distribution is more likely to generate structures with communities. Therefore, angular constraints in the diffusion process are crucial for preserving community structure information. }

\label{fig:GEO}
\end{figure*}

\subsection{Radial Popularity Coordinates. }
In the polar coordinates of hyperbolic space, the radial coordinates of the nodes determine the distortion of the Gaussian distribution.
As shown in Figure \ref{fig:GEO}. (a), the probability density of the central region is lower, while the probability density of the edge region is higher. 
It intuitively follows the laws of physics, the phenomenon of popularity is consistently governed by the rich club effects.
We follow existing hyperbolic methods~\cite{mathieu2019poincare,grattarola2019adversarial,nagano2019differentiable} and use the \textit{wrapped normal distribution} to constructive Gaussian processes in hyperbolic spaces.

\subsection{Angular Similarity Coordinates. }
The angular similarity measures the local connectivity of nodes and is closely related to the community structure of graphs.
As shown in Figure~\ref{fig:GEO}. (b), some existing works~\cite{muscoloni2018npso} couples the latent hyperbolic network geometry to this geometry to generate networks with strong clustering, scale-free degree distribution and a non-trivial community structure. 
Note that our method does not independently introduce noise to the angular similarity metric, but rather incorporates angular geometric constraints into a Gaussian random process to achieve a similar anisotropic effect.

\section{Final Probability Density}\label{app:folded}
First, we derive the form of the probability distribution of the forward diffusion process.

\textbf{Definition: The folded normal distribution.}
If the probability distribution of  $Y$ follows the Gaussian distribution, with $Y{\sim}N\left(\mu,\sigma^2\right)$.
Thus, $X=|Y|$, satisfies $X\sim FN\left(\mu,\sigma^{2}\right)$, where $FN\left(\mu,\sigma^{2}\right)$ denotes the folded normal distribution with mean $\mu$ and variance $\sigma$. The density of $X$ is given by 
\begin{equation}
\begin{aligned}
f\left(x\right)=\frac{1}{\sqrt{2\pi\sigma^2}}\left[e^{-\frac{1}{2\sigma^2}\left(x-\mu\right)^2}+e^{-\frac{1}{2\sigma^2}\left(x+\mu\right)^2}\right]
\end{aligned}
\end{equation}
The density can be written in a more attractive form
\begin{equation}
\begin{aligned}
f\left(x\right)=\sqrt{\frac{2}{\pi\sigma^{2}}}e^{-\frac{\left(x^{2}+\mu^{2}\right)^{2}}{2\sigma^{2}}}\cosh\left(\frac{\mu x}{\sigma^{2}}\right).
\end{aligned}
\end{equation}
Specifically, when $\mu=0$, the density can be represented by 
\begin{equation}
\begin{aligned}
f\left(x\right)=\sqrt\frac{2}{{\pi\sigma^{2}}}e^{-\frac{1}{2\sigma^{2}}x^2}.
\end{aligned}
\end{equation}
which is also named the half-normal distribution.

\textbf{Definition} The random variable $X$ obeys the probability distribution $\mathcal{N_\mathit{f}}\left(\mu,\sigma\right)$ if and only if the density of $X$ can be given by 
\begin{equation}
\begin{aligned}
\label{eqfold}
f(x)=\begin{cases}\sqrt{\frac{2}{\pi\sigma^2}}e^{-\frac{(x-\mu)^{2}}{2\sigma^2}},&x\geq\mu\\0,&x<\mu\end{cases}.
\end{aligned}
\end{equation}

Now we can prove the Theorem \ref{thm:normal}.
\begin{proof}
The angle-constrained noise $z$ in the forward diffusion process is given by Eq \eqref{eq:constrain}. For the convenience of the later derivation, it can be assumed that 
$\mathrm{sgn}\left(\mathrm{logmap}_{\mathbf{o}}^{c}\left(\mathbf{h}_{m}\right)\right)=1$.Thus, according to the definition of the folded normal distribution, it follows that:
\begin{equation}\label{eq:zdist}
\begin{aligned}
z\sim FN\left(0,I\right),\\
\end{aligned}
\end{equation}

Similarly, it can be easily obtained from Eq~\eqref{diff} and Eq~\eqref{eq:zdist} that  the density of $x_t$ in the diffusion process can be written by:
\begin{equation}
\begin{aligned}
f(x_t)=\begin{cases}\sqrt{\frac{2}{\pi\sigma_t^2}}e^{-\frac{(x-\mu_tx_{0})^{2}}{2\sigma_t^2}},&x\geq\mu_tx_0\\0,&x<\mu\end{cases}.
\end{aligned}
\end{equation}
where $\mu_t=\sqrt{\overline{\alpha_{t}} } +\delta\tanh[\sqrt{c}\lambda_{o}^{c} (t)/T_{0} ]$, and $\sigma_t=\left(1-\overline{\alpha _{t}}\right)I$.

Thus, the probability density distribution of $x_t$ satisfies:
\begin{equation}
\begin{aligned}
p(\mathbf{x}_t\mid\mathbf{x}_0)=\mathcal{N_\mathit{f}}\left(\mu_t,\sigma_t\right).
\end{aligned}
\end{equation}
\begin{equation}
\begin{aligned}
\lim_{t \to \infty} \mathbf{x}_{t}\sim \mathcal{N_\mathit{f}}\left(\delta\mathbf{x}_{0} ,I\right).
\end{aligned}
\end{equation}
\end{proof}
\section{Diffusion}\label{app:diffusion details}
\subsection{Loss Prove}
In DDPM, the loss function of the training network is derived from the following equation:
\begin{equation}
\begin{aligned}
\mathbb{E}_q\left[D_{\mathrm{KL}}(p(\mathbf{x}_T|\mathbf{x}_0)\parallel p(\mathbf{x}_T))+\sum_{t>1}D_{\mathrm{KL}}(p(\mathbf{x}_{t-1}|\mathbf{x}_t,\mathbf{x}_0)\parallel p_{\theta}(\mathbf{x}_{t-1}|\mathbf{x}_t))-\log p_{\theta}(\mathbf{x}_0|\mathbf{x}_1)\right].
\end{aligned}
\end{equation}
\begin{proof}
To re-prove the validity of the loss function, it is first necessary to derive the probability distribution of the denoising process according to the Bayesian formulation:
\begin{equation}
\begin{aligned}
p\left(\boldsymbol{x}_{t-1}|\boldsymbol{x}_t,\boldsymbol{x}_0\right)& =\frac{p\left(\boldsymbol{x}_t|\boldsymbol{x}_{t-1},\boldsymbol{x}_0\right)p\left(\boldsymbol{x}_{t-1}|\boldsymbol{x}_0\right)}{p\left(\boldsymbol{x}_t|\boldsymbol{x}_0\right)}  \\
&\propto exp[-\frac{(x_{t-1}-\mu_{t-1}x_0)^{2}}{2(1-\overline{\alpha}_{t-1})}-\frac{(x_t-(\sqrt{\alpha_t}x_{t-1}+\beta_tx_0)]^{2}}{4(1-\alpha_t)}+\frac{(x_t-\mu_t x_0)^{2}}{2(1-\overline{\alpha}_t)}]\\
&=exp\left\{-\frac{1}{2} [\frac{(x_{t-1}-\mu_{t-1}x_0)^{2}}{(1-\overline{\alpha}_{t-1})}+\frac{(x_t-(\sqrt{\alpha_t}x_{t-1}+\beta_tx_0)]^{2}}{2(1-\alpha_t)}-\frac{(x_t-\mu_t x_0)^{2}}{(1-\overline{\alpha}_t)}]\right\}\\
&=exp\left\{-\frac{1}{2} [(\frac{1}{1-\overline{\alpha}_{t-1}}+\frac{\alpha_t}{2(1-\alpha_t)})x_{t-1}^2-(\frac{2\mu_{t-1}x_0}{1-\overline{\alpha}_{t-1}}+\frac{2\sqrt{\alpha_t}(x_{t}-\beta_tx_0)}{(1-\alpha_t)})x_{t-1}+C(x_{t},x_0)]\right\}\\
&\propto \mathcal{N_\mathit{f}}\left(\mu_q,\sigma_q\right).
\end{aligned}
\end{equation}
where $\mu_q$ is given by:
\begin{equation}
\begin{aligned}
\mu_q=\frac{2[\mu_{t-1}(1-\alpha_t)x_0+(1-\overline{\alpha}_{t-1})\sqrt{\alpha_t}(x_{t}-\beta_tx_0)]}{2-\alpha _t-\overline{\alpha }_t } 
\end{aligned}
\end{equation}

Based on the above derivation, our goal translates to:
\begin{equation}
\begin{aligned}
&\underset{\theta}{\operatorname*{\arg\min}}D_{\mathrm{KL}}(q(\boldsymbol{x}_{t-1}|\boldsymbol{x}_t,\boldsymbol{x}_0)\parallel p_{\boldsymbol{\theta}}(\boldsymbol{x}_{t-1}|\boldsymbol{x}_t))\\
&=\underset{\theta}{\operatorname*{\arg\min}}D_{\mathrm{KL}}(\mathcal{N_\mathit{f}}(\boldsymbol{x}_{t-1};\boldsymbol{\mu}_q,\boldsymbol{\Sigma}_q(t))\parallel\mathcal{N}(\boldsymbol{x}_{t-1};\boldsymbol{\mu}_{\boldsymbol{\theta}},\boldsymbol{\Sigma}_q(t)))\\
&=\underset{\theta}{\operatorname*{\arg\min}}\int_0^\infty\sqrt{\frac{2}{\pi\sigma_q^2}}\left[e^{-\frac{1}{2\sigma_q^2}(x-\mu_q)^2}\right]\log\frac{\sqrt{\frac{2}{\pi\sigma_q^2}}\left[e^{-\frac{1}{2\sigma^2}(x-\mu_q)^2}\right]}{\sqrt{\frac{2}{\pi\sigma^2}}e^{-\frac{1}{2\sigma^2}(x-\mu_\theta)^2}}dx\\
&=\underset{\theta}{\operatorname*{\arg\min}}\int_0^\infty\sqrt{\frac{2}{\pi\sigma_q^2}}\left[e^{-\frac{1}{2\sigma_q^2}(x-\mu_q)^2}\right] [-\frac{1}{2\sigma^2}(x-\mu_q)^2+\frac{1}{2\sigma^2}(x-\mu_\theta)^2]log_{2}edx\\
&=\underset{\theta}{\operatorname*{\arg\min}}\int_0^\infty\left|(x-\mu_q)^2-|(x-\mu_\theta)^2\right|dx\\
&=\underset{\theta}{\operatorname*{\arg\min}}\left|(\mu_\theta-\mu_q)^2\right|
\end{aligned}
\end{equation}
Thus after a derivation similar to that of DDPM, we can define the loss function as:
\begin{equation}
\begin{aligned}
\mathcal{L}=\mathbb{E}\left\|f_{\theta}\left(X_{t},A,t\right)-X_{0}\right\|^{2}.
\end{aligned}
\end{equation}
\end{proof}
\subsection{Sample}
As demonstrated in Theorem \ref{thm:k}, the individual tangent planes can be viewed as the tangent planes of a hybrid surface. The samples can also be sampled directly in the same tangent plane and denoised uniformly. 
Sampling from the target distribution is difficult, but sampling from its construction process is easy. We just need to compute an additional direction matrix. The direction transformation matrix is given by:
\begin{equation}
\begin{aligned}\label{direction}
T=\mathrm{sgn}\left(\mathrm{logmap}_{\mathbf{o}}^{c}\left(\mathbf{h}_{m_1}\right),\mathrm{logmap}_{\mathbf{o}}^{c}\left(\mathbf{h}_{m_2}\right),...,\mathrm{logmap}_{\mathbf{o}}^{c}\left(\mathbf{h}_{m_i}\right)\right)
\end{aligned}
\end{equation}

\subsection{Reverse Denosing Process}
In this section, we follow the same denoising process as in DDM \cite{yang2023directional} with the following algorithm. Specifically, the denoising process is given by:
\begin{equation}\label{denoise}
\begin{aligned}
&\hat{x}_0 = R(x_s,s)\\&x_{s-1}=x_s-D(\hat{x}_0,s)+D(\hat{x}_0,s-1)
\end{aligned}
\end{equation}

where $R$ is the restoration operator, $D$ is the addition noise operator, $s$ represents the time steps.

Next, we prove that this algorithm still satisfies our diffusion process:
\begin{equation}
\begin{aligned}
&z(x_{t},t)=\frac{x_{t}-(\sqrt{\alpha_{t}}+\delta\tanh[\sqrt{c}\lambda_{o}^{c} (t)/T_{0} ])\hat{x_{0}}}{\sqrt{1-\alpha_{t}}},
\end{aligned}
\end{equation}
\begin{equation}
\begin{aligned}
&D(\hat{x_0},t)=(\sqrt{\alpha_t}+\delta\tanh[\sqrt{c}\lambda_{o}^{c} (t)/T_{0} ])\hat{x_0}+\sqrt{1-\alpha_t}\hat{z},
\end{aligned}
\end{equation}

Thus, the denoising process can obtain $x_{t-1}$:
\begin{equation}
\begin{aligned}
x_{t-1} =&x_{t}-D(\hat{x_{0}},t)+D(\hat{x_{0}},t-1)  \\
=&x_{t}-[(\sqrt{\alpha_t}+\delta\tanh[\sqrt{c}\lambda_{o}^{c} (t)/T_{0} ])\hat{x_0}+\sqrt{1-\alpha_t}\hat{z}]+[(\sqrt{\alpha_{t-1}}+\\&\delta\tanh[\sqrt{c}\lambda_{o}^{c} (t-1)/T_{0} ])\hat{x_0}+\sqrt{1-\alpha_{t-1}}\hat{z}] \\
=&(\sqrt{\alpha_{t-1}}+\delta\tanh[\sqrt{c}\lambda_{o}^{c} (t-1)/T_{0} ])\hat{x_0}+\sqrt{1-\alpha_{t-1}}\hat{z}
\end{aligned}
\end{equation}
The complete denoising process algorithm is described in Algorithm \ref{Alg:sampling}.

\begin{algorithm}[tb]
    \caption{Sampling from \modelname} 
    \label{Alg:sampling}
\begin{algorithmic}
    \STATE {\bfseries Input:} Graph $\mathcal{G}=\{\mathbf{A},\mathbf{X}\}$; Number of diffusion timesteps $T$;
    \STATE {\bfseries Output:} The generated hyperbolic embedding coordinates $\hat{x}_\mathbb{H}$.
    \STATE sample $x_{T}$ from the folded normal distribution $\gets$ Eq.~\eqref{direction}  \\
    \FOR{$t=T$ {\bfseries to} $0$}
        \STATE predict $\hat{x}_{0}$ with $\hat{x}_{t}$,$t$ and $A$;\\
        \STATE de-noise and predict $\hat{x}_{t-1}$  $\gets $ Eq.~\eqref{denoise};\\
    \ENDFOR
    \STATE mapping the generated $\hat{x}_{0}$ to the Poincar\'e embedding  $\hat{x}_\mathbb{H}$
\end{algorithmic}  
\end{algorithm}

\section{Complexity Analysis}\label{app:complexity}
In this part, we will analyze the time complexity and space complexity of our algorithm using graph set data as an example.   $G=\left ( X,E \right )$  is one of the graphs set $G^s$, where $X$ is the $n$-dimensional node eigenvector and  $E$ is the $m*m$-dimensional adjacency matrix of the graph.  $s$ is the number of graphs in the graph set $G^s$.
\subsection{Time Complexity}
\textbf{Hyperbolic embedding and clustering:} First, we need to embed each graph $G$ into a hyperbolic space by HGCN. Let the graph  $G$  be embedded after passing through HGCN into the  $m*d$-dimensional vector $H^m$. The time complexity of this process can be viewed as $O(md)*1(t)$, where $1(t)$ denotes the time through the neural network. The process of clustering can be approximated with a time complexity of  $O(kmd)$, where k denotes the number of clusters. Both the embedding process and the clustering process occupy a very short time compared to the forward diffusion and training denoising network process.

\textbf{Diffusion-forward process:}For the forward diffusion process, we need to adjust the noise direction according to the north pole to the center of the mass vector direction. Since calculating the direction transfer matrix can be prepared in advance before training, the complexity of this part is $O(md)$. The rest of the process is the same as the normal diffusion process, and it is sufficient to do the noise addition process once. The time complexity of this part is $O(md)$.

\textbf{Training of denoising networks:} The training of denoising networks is essentially the same as other diffusion models and does not require additional computing time. So, the time complexity of this part is $O(md)*1(t)$

Overall, the time complexity of the diffusion process is $O(1(t)*2md)+O((k+2)md)$ in one epoch.

\textbf{Generation:} The main difference between our generation process and other methods is in our noise sampling. Since we need to sample in different tangent planes in hyperbolic space, which requires each time to judge the tangent plane where the corresponding sampling point is located, the time complexity of this process is O(m). Compared to the subsequent stepwise denoising process, the sampling process takes very little time, which means that we use almost the same amount of time as the other methods for generation.

\subsection{Space Complexity}
It is worth noting that compared to other graph diffusion models, our model has a significant advantage over other models due to its space complexity, which indirectly leads to a shorter training time than other models.
Specifically, in other graph diffusion models, such as GDSS and Digress, since it is directly for the sparse adjacency matrix diffusion, then as to the graph set $G^s$, which is equivalent to the diffusion scale of $O(s \times m^2 \times d)$, the number of parameters that need to be trained for the denoising network is also namely huge.

However, in our approach, since we embed the graphs in hyperbolic space, each graph is represented as a $m*d$-dimensional vector in the hyperbolic space, which means that our diffusion scale is $O(sm ^2d)$, and the corresponding number of parameters in the training network is much smaller than in other approaches (For large-scale graphs, m is much larger than $d$). This suggests that our approach has significant advantages in both time and space.

\section{Datesets Description}\label{app:datasets}
We show the specific properties of the dataset in Table~\ref{dataset_description}.
\begin{table}[t]\label{tab:real}
\caption{Statistics of real-world datasets.}
\small
\centering
\begin{tabular}{cl|ccccc}
\toprule
\multicolumn{2}{c|}{\textbf{Dataset}}  & \textbf{\#Nodes}& \textbf{\#Edges}  & \textbf{\#Features}&\textbf{\#Avg. Degree} & \textbf{\#Class}  \\ 
\midrule
\multirow{3}{*}{\rotatebox{90}{\textbf{Link P.}}}  
&\textbf{Cora}     & 2,708  & 5,429 &1,433 & 3.90  & 7        \\
&\textbf{Citeseer}  & 3,312 & 4,732 &3,703 & 2.79  & 6       \\
&\textbf{Polblogs}    & 1,490  & 19,025  &500 & 25.54  & 3     \\
\midrule
& \textbf{Dataset}    & \textbf{\#Graphs} & \textbf{\#Avg. Node} & \textbf{\#Avg. Edge} &\textbf{\#Max Num Node} & \textbf{\#Class} \\ 
\midrule
\multirow{4}{*}{\rotatebox{90}{\textbf{Graph G.}}}  
&\textbf{MUTAG}     & 188 & 17.9 & 39.6 & 28   &2     \\
&\textbf{IMDB-B}  & 1,000 & 19.8 & 193.1  & 136  &2       \\
&\textbf{PROTEINS}    & 1,113 & 39.1 & 145.6  & 620  &2   \\
&\textbf{COLLAB}    & 5,000  & 74.5 & 4914.4   & 492  &3  \\
\bottomrule
\end{tabular}
\label{dataset_description}
\end{table}


  

\section{Sensitivity Experiment}\label{app:senstivity}

\subsection{$k$-cluster}
To test the effect of different numbers of clusters on the diffusion effect, we tested it on three datasets with different geometrical structures: Cora, SBM and Hierarchical Fractal. SBM portrays five equally partitioned communities with the edge creation of intra-community $p$ =0.21 and intra-community $q$ - 0.025 probabilities. In total, it consists of 13000 nodes with 2000,20000,3000,3000,3000 nodes in each category respectively. Hierarchical Fractal is a pure tree-like self-similar structure with three edges for each node respectively. Experimental settings: $\delta$=0.4, $T_{0}$=1000, the epoch in diffusion training = 10000, the optimizer uses Adam and the learning rate is 0.0001.

The results of the experiment are depicted in the figure, where three datasets are analyzed: Cora, SBM, and Hierarchical Fractal.
(1)Cora: Being a real dataset, Cora exhibits a highly complex structure. A larger value of k can better capture its data characteristics. However, excessively large values of k, such as k=500, introduce significant anisotropy, where, on average, only 4 nodes belong to a class. This excessive anisotropy poses challenges in learning the dataset's probability distribution.
(2)SBM: Despite having only five major categories, SBM displays a complex structure within each community. Increasing k yields improved results, likely due to the intricacies within each community that benefit from a higher resolution.
(3)Hierarchical Fractal: Interestingly, changes in k have minimal impact on the training effectiveness of Hierarchical Fractal. This phenomenon may stem from the self-similarity inherent in its structure. Unlike other datasets, where clustering involves deconstructing mixed manifolds, fractal data maintains consistent local geometric properties due to its self-similar structure.

\subsection{Priori Geometry}
To explore the ability of geometric prior information to enhance the effect of diffusion, we further did a sensitivity analysis on $\delta$ to explore the performance of the model on the NC task for $\delta$=0.2,0.4,0.6,0.8,1.0, respectively. A uniform setting of k = 200 was used, and the rest of the experimental settings were kept consistent with the experiments on clustering sensitivity.

The experiment results depicted in the figure illustrate a progressive improvement as the parameter $\delta$ gradually increases from very small values. This improvement is attributed to the retention of more geometric a priori information during the diffusion process. However, as $\delta$ continues to increase, there is an initial decline in effectiveness, possibly due to the heightened geometric prior information. Actually, the increase in geometric information also leads to an increase in anisotropy during diffusion, resulting in a reduced variance among nodes within the same tangent space. Consequently, learning an optimal probability distribution becomes more challenging. As the geometric prior information further increases, a rebound effect is observed, indicating that the increment in prior information surpasses the interference caused by anisotropy on node representation. 

\section{Additional Results}\label{app:exper results}
In this section, we record the results for the datasets beyond the in-text presentations. 
\subsection{MMD Scores of Graph Generation Task}
The results of the MMD metrics~\cite{GraphRNN2018} for the graph generation task are reported in Table~\ref{table:synthetic_mmd2}, and we obtain conclusions consistent with those described above. 
Our MMD is computed with the RBF kernel which is a more stable and comprehensive metric to measure the diversity and realism of generated graphs.
\vspace{-0.5cm}
\begin{table*}[htbp]
\caption{Generation additional results about the MMD distance between the original and generated graphs.}
\centering
\resizebox{\textwidth}{!}{
\begin{tabular}{c|ccc|ccc|ccc|ccc}
\toprule
\multirow{3}{*}{\textbf{Method}} & \multicolumn{6}{c|}{\textbf{Synthetic Datasets}}    & \multicolumn{6}{c}{\textbf{Real-world Datasets}}  \\ 
\cline{2-13} 

& \multicolumn{3}{c}{\textbf{Ego}} & \multicolumn{3}{c|}{\textbf{Grid}}    & \multicolumn{3}{c}{\textbf{IMDB-B}}    & \multicolumn{3}{c}{\textbf{COLLAB}}  \\ 
\cline{2-13} 
                        & Degree                & Cluster   & Spectre             & Degree    & Cluster            & Spectre   & Degree        & Cluster        & Spectre   & Degree        & Cluster       & Spectre  \\ 
\midrule
VGAE    & 0.414 & 0.156 & 0.456        & \textbf{0.050}    & 2.000    & 0.145        & 0.514   & 1.405     & 0.700  & 0.418 & 1.174 &0.700  \\ 
GraphRNN   & 0.206 & 0.539  &  0.157       & 0.203   & \underline{0.043}   & \textbf{0.042}        & 0.137  & \textbf{0.252}    & 0.423  & 0.044  & 0.036 & 0.510 \\ 
\midrule
GDSS    & 1.034 & 0.143 & 0.667       & \underline{0.111}   & \textbf{0.005} & 0.886        & 0.904   &1.729 & 0.748   & 0.773  & 1.589 & 0.502 \\ 
DiGress   & 0.110  & \underline{0.056} &  0.122       &  0.689  & 1.115  &  0.203       & 0.166   & 0.425    & 0.159   & \underline{0.022}  & \textbf{0.008} & \textbf{0.003} \\ 
GraphGDP   & 0.059 & 0.115 & 0.054        & 0.872   & 1.001  & 0.174        & 0.123   & 0.638    & 0.257   & 0.092 & 0.291 & 0.138 \\ 
EDGE   & \textbf{0.023} & \textbf{0.048} &\underline{ 0.023}       &  0.223  & 0.072  &  0.802       & \underline{0.041}   & 0.874    & \textbf{0.026}  & 0.023 & 0.569 & 0.034\\ 
\midrule
\textbf{\modelname}    & \underline{0.075} & 0.090
 &\textbf{ 0.015}        & 0.422   & 0.665
  & \underline{0.137}        & \textbf{0.034}   & \underline{0.257}    & \underline{0.030}  & \textbf{0.016}  & \underline{0.023}  & \underline{0.009} \\ 
\bottomrule
\end{tabular}}
\label{table:synthetic_mmd2}
\end{table*}
\vspace{-0.5cm}


\subsection{F1 Scores of Graph Generation Task}
The results of the F1 pr and F1 dc metrics~\cite{F1-prdc} for the graph generation task are presented in Table~\ref{table:f1_prdc}, and we obtain conclusions consistent with those described above where larger F1 means the model has better fidelity and diversity. 
Note that F1 pr is the harmonic mean of improved precision and recall and
F1 dc is the harmonic mean of density and coverage.
They are sensitive to detecting mode collapse and mode dropping.

\begin{table*}[htbp]
\caption{Generation results about the F1 score of precision-recall and density-coverage (F1 pr and F1 dc) between the original and generated graphs.}
\centering
\resizebox{\linewidth}{!}{
\begin{tabular}{c|cccccccc|cccccccc}
\toprule
\multirow{3}{*}{\textbf{Method}} & \multicolumn{8}{c|}{\textbf{Synthetic Datasets}}   & \multicolumn{8}{c}{\textbf{Real-world Datasets}}    \\ 
\cline{2-17} 

& \multicolumn{2}{c}{\textbf{Community}} & \multicolumn{2}{c}{\textbf{BA-G}}    & \multicolumn{2}{c}{\textbf{Ego}}    & \multicolumn{2}{c|}{\textbf{Grid}} & \multicolumn{2}{c}{\textbf{MUTAG}} & \multicolumn{2}{c}{\textbf{PROTEINS}}    & \multicolumn{2}{c}{\textbf{IMDB-B}}    & \multicolumn{2}{c}{\textbf{COLLAB}}  \\ 
\cline{2-17} 
                        & F1 pr  & F1 dc  & F1 pr  & F1 dc  & F1 pr  & F1 dc  & F1 pr  & F1 dc  & F1 pr  & F1 dc  & F1 pr  & F1 dc  & F1 pr  & F1 dc  & F1 pr  & F1 dc   \\ 
\midrule
VGAE    & 0.001 & 0.023 & 0.104 & 0.002 & 0.162 & 0.139& 0.091 & 0.080 &0.173 & 0.072 &  0.064 & 0.204 &0.031&0.165 &0.283&0.173\\ 
GraphRNN   &0.732 &\textbf{1.312} & 0.016 & 0.003 & 0.008 & 0.186 & 0.333 & 0.109 & 0.643 & 0.334 & 0.804 & 0.842 & 0.475 & 0.575 &0.260&0.295 \\ 
\midrule
GDSS    & 0.245 & 0.157 & 0.020 & 0.132  & 0.201 & 0.093 & 0.496 & 0.260 & 0.585 & 0.428  & 0.763 & 0.730 & 0.048 & 0.073 & 0.019  & 0.005\\ 
DiGress  & 0.645 & 0.695 & 0.724 & 0.783 & 0.409 & 0.091 & 0.761 & \textbf{0.742} & 0.713 & 0.549  & 0.867 & 0.193 &0.546 &0.358 & 0.918 & \textbf{1.179} \\ 
GraphGDP  & \underline{0.815}  & 0.960 &  0.157 & 0.060 & \underline{0.933} &\textbf{0.997} & \textbf{0.815} & 0.627  & \textbf{0.880} & \textbf{0.838} &0.921 &0.931 & 0.027 & 0.015 & \textbf{0.989} & \underline{1.016} \\ 
EDGE   & \textbf{0.989}  & \underline{1.207} & \underline{0.920} & \underline{0.864} & 0.609 & \underline{0.770} & 0.671 & 0.287 &0.667 & 0.679 & \underline{0.951} & \textbf{0.974} & \textbf{0.959} & \textbf{0.995} & 0.933 & 1.015 \\ 
\midrule
\textbf{\modelname}   &  0.812 & 0.864 & \textbf{0.971} & \textbf{0.923} & \textbf{0.943} & 0.734 & \underline{0.792} & \underline{0.641}  & \underline{0.730} & \underline{0.680} &\textbf{0.973} & \underline{0.943} & \underline{0.730} & \underline{0.670} & \underline{0.981} & 0.653\\ 
\bottomrule
\end{tabular}}
\label{table:f1_prdc}
\end{table*}

\subsection{Visualization}\label{app:visualization}
\begin{figure*}[h]
\centering
\includegraphics[width=0.859\textwidth]{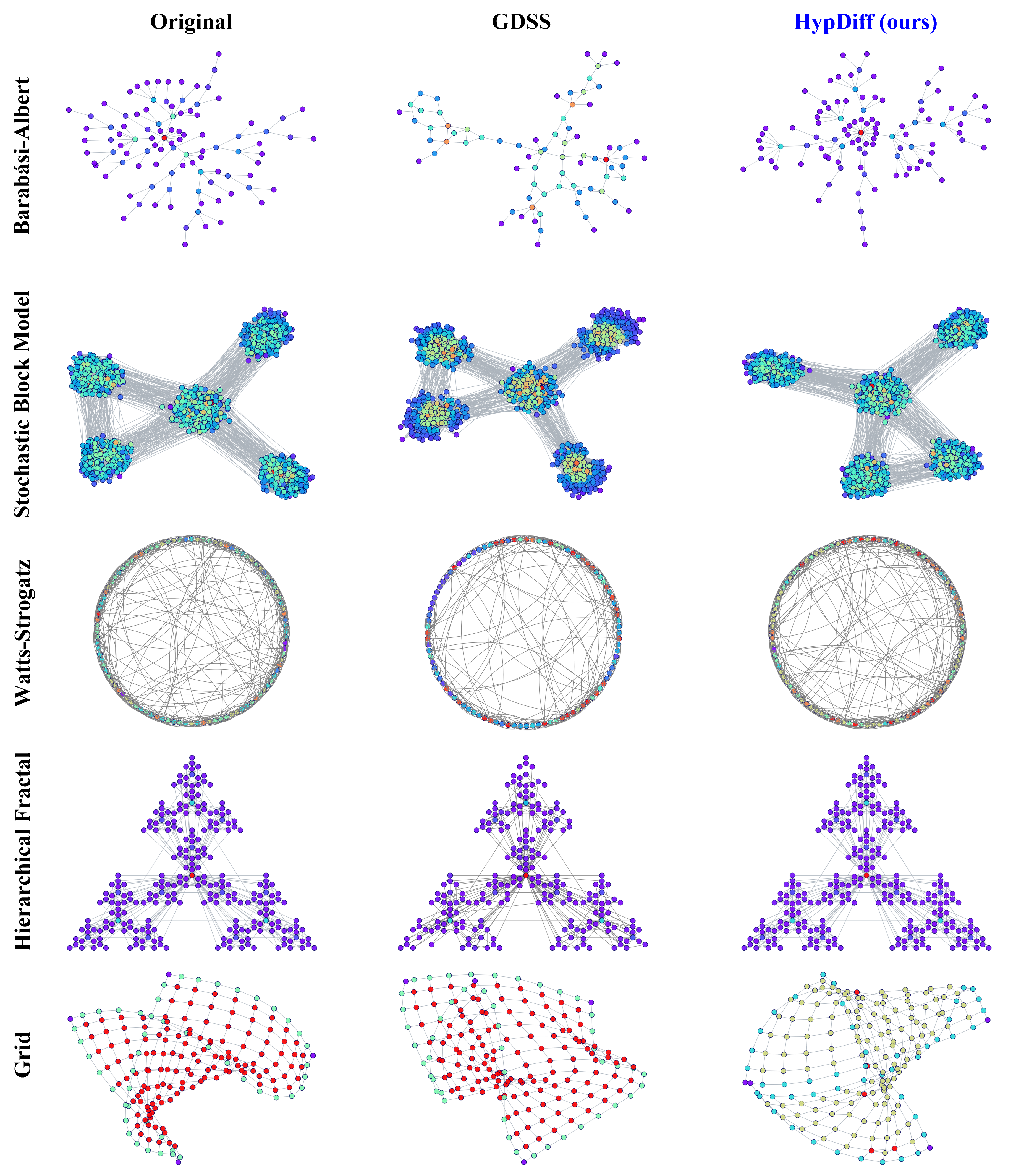} 
\caption{Visualization of graph generations.}
\label{fig:visualization}
\end{figure*}

The visualization of~\modelname~and GDSS by five well-accepted graph theoretical models is shown in Figure~\ref{fig:visualization}
These graphs can represent typical general and complex topological properties:
\textbf{BA}~\cite{barabasi1999emergence} scale-free graphs have more tree structure. \textbf{SBM}~\cite{holland1983stochastic} graphs have more community structure. \textbf{WS}~\cite{watts1998collective} small-world graphs have more cyclic and clique structure.  \textbf{Hierarchical Fractal} ~\cite{ravasz2003hierarchical} networks is constructed by self-organization and self-similarity. \textbf{Grid}~\cite{GraphVAE} have regular (Euclidean) structures.
While comparing the generated structure, we also color the node by using its degree. 
The results demonstrate that our ~\modelname~ exhibits significantly enhanced proficiency in reproducing the original graph structure across BA, SBM, WS, and Fractal graphs, while consistently achieving a coherent distribution of node colors. 
Regarding the Grid, due to the weakness of hyperbolic space for capturing regular structure, \modelname~still generates nodes with high degrees (the red color distribution is uneven).

\section{Analysis of GPU Occupancy and Runtime}\label{gpu}
We conducted a statistical analysis of GPU utilization during the model training and the average denoising time over 1000 steps during diffusion. The results, presented in Table~\ref{table:time and gpu}, indicate that our model performs comparably in terms of time to other models, yet it utilizes significantly less GPU. This implies that our model exhibits higher efficiency.
\begin{table*}[htbp]
\caption{GPU memory usage during training and time used to denoise 1000 steps during generation.
}
\centering
\resizebox{\linewidth}{!}{
\begin{tabular}{c|cccccc|cccccc}
\toprule
\multirow{3}{*}{\textbf{Method}} & \multicolumn{6}{c|}{\textbf{Synthetic Datasets}}   & \multicolumn{6}{c}{\textbf{Real-world Datasets}}    \\ 
\cline{2-13} 

& \multicolumn{2}{c}{\textbf{Community}} & \multicolumn{2}{c}{\textbf{BA-G}}    & \multicolumn{2}{c}{\textbf{Ego}}     & \multicolumn{2}{c}{\textbf{MUTAG}} & \multicolumn{2}{c}{\textbf{PROTEINS}}    & \multicolumn{2}{c}{\textbf{IMDB-B}} \\ 
\cline{2-13} 
                        & Time(s)  & GPU(MB)  & Time(s)  & GPU(MB)  & Time(s)  & GPU(MB) &Time(s)  & GPU(MB)  & Time(s)  & GPU(MB)  & Time(s) & GPU(MB)    \\ 
\midrule
GDSS    & 10.14 & 3475 & 11.80 & 7750  & 9.71 & 3883  & 10.79 & 3907  & 12.04 & 6305 & 12.5 & 3501 \\ 
DiGress  & 9.62 & 3936 & 11.42 & 9012  & 12.28 & 4174 & 9.84 & 4125  & 11.74 & 6975 &12.1 &5800  \\ 
GraphGDP  & 12.58 & 3802 &  14.36 & 13164 & 12.47 &3848   & 12.85 & 3956 &12.18 &44708 & 13.6 & 5902  \\ 
EDGE   & 9.87  & 2825 & 11.83 & 25236 & 10.24 &2657   &9.73 &27603 & 10.95 & 26188 & 11.8 & 6205  \\ 
\midrule
\textbf{\modelname}   &   10.03 & 2246 & 12.04 & 5697 & 10.15 & 2570   & 9.92 & 2720 &10.72 &4735 & 11.20 & 2519 \\ 
\bottomrule
\end{tabular}}
\label{table:time and gpu}
\end{table*}


\end{document}